\documentclass[review]{elsarticle}   % Exact Times New Roman (XeLaTeX); spacing is set explicitly below
\usepackage{amsmath,amssymb,amsthm}
\usepackage{newtxtext}
\usepackage{newtxmath}
\usepackage{xcolor}
\usepackage{graphicx}
\usepackage{booktabs}
\usepackage{multirow}
\usepackage{algorithm}
\usepackage{algpseudocode}
\usepackage[hidelinks]{hyperref}
\usepackage{tikz}
\usetikzlibrary{positioning,arrows.meta}
\usepackage[top=4.3cm,bottom=4.3cm,left=4.8cm,right=4.8cm]{geometry}
\usepackage{setspace}
\usepackage{etoolbox}

\makeatletter
\let\@blstr\baselinestretch
\patchcmd{\abstract}{\def\baselinestretch{1}}{\doublespacing}{}%
  {\PackageError{danco-submission}{Abstract spacing patch failed}{Check elsarticle.cls}}
\patchcmd{\keyword}{\normalsize\normalfont\def\baselinestretch{1}}%
  {\def\baselinestretch{1}\normalsize\normalfont}{}%
  {\PackageError{danco-submission}{Keyword spacing patch failed}{Check elsarticle.cls}}
\patchcmd{\pprintMaketitle}{\Large\@title}%
  {\fontsize{14}{17}\selectfont\@title}{}%
  {\PackageError{danco-submission}{Title font patch failed}{Check elsarticle.cls}}
\patchcmd{\pprintMaketitle}{\normalsize\elsauthors}%
  {\fontsize{8}{10}\selectfont\elsauthors}{}%
  {\PackageError{danco-submission}{Author font patch failed}{Check elsarticle.cls}}
\newcommand{\prtitlegap}[2]{%
  \patchcmd{\pprintMaketitle}{\vskip#1pt}{\vskip#2pt}{}%
    {\PackageError{danco-submission}{Title spacing patch failed}{Check elsarticle.cls}}}
\prtitlegap{18}{6}
\prtitlegap{36}{4}
\prtitlegap{10}{4}
\prtitlegap{10}{4}
\prtitlegap{10}{4}
\prtitlegap{12}{4}
\prtitlegap{12}{4}
\def\ps@pprintTitle{%
  \let\@oddhead\@empty\let\@evenhead\@empty
  \def\@oddfoot{\hfil\thepage\hfil}\let\@evenfoot\@oddfoot}
\makeatother
\AfterEndEnvironment{frontmatter}{\doublespacing}

\newtheorem{proposition}{Proposition}
\newtheorem{lemma}{Lemma}

\newcommand{\dhat}{\hat{d}}
\newcommand{\KL}{\mathrm{KL}}
\newcommand{\atantwo}{\operatorname{atan2}}
\DeclareMathOperator*{\argmin}{arg\,min}

\journal{Pattern Recognition}

\begin{document}

\begin{frontmatter}

\title{Interpretable intrinsic dimension estimation through componentwise
       calibration of distance and angle}

\author[nsysu]{Chih-Hsuan Huang}
\author[nsysu]{Chih-Wei Chen}
\author[nsysu]{Szu-Chi Chung\corref{cor1}}
\cortext[cor1]{Corresponding author.}
\ead{phonchi@math.nsysu.edu.tw}

\affiliation[nsysu]{organization={Department of Applied Mathematics, National Sun Yat-sen University},
                    addressline={No. 70, Lienhai Rd.},
                    city={Kaohsiung}, postcode={80424}, country={Taiwan}}

\begin{abstract}
DANCo (Dimensionality from Angle and Norm Concentration) jointly calibrates
nearest-neighbor distance and angular statistics and consistently reaches
state-of-the-art accuracy on clean intrinsic-dimension (ID) benchmarks.
Practical data, however, introduce neighborhood-relative noise and
sample-amplitude heterogeneity that can distort these geometric signals. We
reformulate DANCo componentwise, retaining separate distance and angular
discrepancy curves so that the source of an estimate can be identified and
interpreted. For the distance component, we derive a closed-form
Kullback--Leibler divergence for the generic-order ratios of the
generalized ratios ID estimator (Gride); when both
angular parameters are matched (Full), Gride reduces mean percentage error
from $27.7\%$ to $17.6\%$ at noise equal to $40\%$ of typical neighbor
spacing on 24 manifolds. For the angular component, two sampling regimes
motivate aligning mean direction while retaining concentration matching
(Profiled). On a Gaussian scale mixture with generating dimension 70
embedded in 100 dimensions, profiling raises the Minimum Neighbor Distance
(MiND) estimate from $22.8$
to $66.7$, while removing the known amplitudes restores MiND--Full to
$71.9$; the control thus attributes the Full shortfall to amplitude
heterogeneity.
On CIFAR-10 and ImageNet, amplitude-reducing normalizations move angular
location toward the references and narrow the Full--Profiled gap, an
observational counterpart to the controlled mixture. Across four
pretrained convolutional neural networks, Gride--Profiled, the
two-nearest-neighbor estimator (TWO-NN), and the maximum-likelihood
estimator (MLE) exhibit similar rise-and-fall profiles, while Full--Profiled
differences identify the layers most sensitive to angular calibration.
\end{abstract}


\end{frontmatter}

% !TEX root = ../main.tex
\section{Introduction}
\label{sec:intro}

Intrinsic dimension (ID) helps characterize dataset complexity, guide
dimensionality reduction, and interpret geometric changes across
neural-network representations~\cite{camastra2016intrinsic,ansuini2019intrinsic}. It is the number of
coordinates needed to represent local variability within a higher-dimensional
observation space. Two local geometric signatures carry information about ID:
nearest-neighbor distances reflect volume growth, while angles between
neighbor displacements reflect angular concentration. Calibrating these
signatures connects the geometry of an observed sample to a dimension estimate.

Dimensionality from Angle and Norm Concentration (DANCo) calibrates both
signatures against simulated references indexed by candidate dimension.
Their finite-sample errors respond differently to neighborhood geometry:
distance-based estimates tend to underestimate the generating dimension,
whereas angle-based estimates tend to overestimate it. The original DANCo
study describes these errors as loosely counterbalancing~\cite{ceruti2014danco}, and subsequent work quantifies angular
overestimation~\cite{thordsen2022abid}. Joint reference matching lets the two
signals support a common candidate and reduces the finite-sample high-ID bias
of purely distance-based estimates. Across independent clean benchmarks, DANCo is consistently among the state of the art: DANCo has the lowest
mean percentage error and leading Friedman rank in the Campadelli et al.
benchmark~\cite{campadelli2015intrinsic}, and independent evaluations place
it first or among the most accurate~\cite{benko2022manifold,qiu2022intrinsic,qiu2023underestimation,binnie2025survey}.
A recent comparison finds its spline-based variant, FastDANCo, most
reliable at high dimension~\cite{tostiguerra2025intrinsic}.
DANCo serves as a state-of-the-art reference in comparisons of local
estimators~\cite{denti2022generalized,facco2017estimating,gomtsyan2019geometry}.

Practical data add three conditions these benchmarks do not stress. First, measurement
and preprocessing noise perturb the smallest neighbor distance first, so
ratios based on larger neighbor orders are more robust to it (Section~\ref{sec:exp-noise}). Second, image
contrast and unnormalized neural activations produce observations with
different centered norms, a condition we call
\emph{sample-amplitude heterogeneity}. Through neighbor selection, this
variation changes both distance and angular statistics even when the
distance-ratio orders are fixed. Third, learned representations require candidate dimensions in the hundreds, where the
thin-shell angular regime makes sample-amplitude effects on angular location
consequential (Section~\ref{sec:method-nustrip}), while ordinary Bessel evaluation of the angular reference
can also overflow (Section~\ref{sec:method-impl}). A combined dimension estimate alone does not
show which component determined its value or how these conditions affected it.

We develop a componentwise formulation that makes those influences explicit
and lets each part of the calibration address its condition. Separate distance,
angular, and combined discrepancy curves reveal agreement between the
geometric signals and identify which term shifts the minimum. For the
distance component, we use the generic-order ratio introduced by
Gride~\cite{denti2022generalized} and derive the Kullback--Leibler (KL)
divergence needed to calibrate its law. For the angular component, two
sampling regimes distinguish concentration from a mean direction that changes
with geometry and sample amplitude. We call matching both von Mises
parameters \emph{Full} and aligning mean direction before matching
concentration \emph{Profiled}; reporting both exposes the effect of angular
location. Stable Bessel evaluation extends the calibrated analysis to
candidate dimensions in the hundreds.

We make four contributions:

\begin{enumerate}
\item We formulate componentwise calibration through explicit statistics,
references, discrepancies, and nuisance-parameter treatments. Separate and
joint minima make the influence of distance and angular geometry
interpretable. The formulation also supports interchangeable components:
a statistic can fill either slot when its dimension-indexed reference law
carries a computable discrepancy
(Section~\ref{sec:method-framework}).

\item We derive a closed-form KL divergence for generic-order Gride ratios.
This brings their robustness to neighborhood-relative noise into the
calibrated objective
(Section~\ref{sec:method-gride}).

\item We distinguish local-interior and high-dimension, low-sample-size angular
limits to motivate mean-direction profiling. A Gaussian scale mixture tests
profiling under controlled sample-amplitude heterogeneity, and dividing out
the known simulated amplitudes recovers Full calibration by removing the
angular-reference mismatch
(Sections~\ref{sec:method-nustrip} and~\ref{sec:exp-gsm}).

\item We carry componentwise analysis into learned representations. Scaled
Bessel evaluation and distance-specific reference surfaces built from a shared
simulation grid make calibrated estimation practical through candidate
dimension $400$. On natural images, the component statistics show how
amplitude-reducing normalization moves angular location toward the reference.
Across four pretrained convolutional neural networks (CNNs), Gride--Profiled
layer-wise profiles share the rise-and-fall course of the two-nearest-neighbor
and maximum-likelihood estimators, and paired Full--Profiled comparisons
place the largest median separation at the Profiled peak, with
Full estimates $100$--$214$ dimensions lower
(Sections~\ref{sec:method-impl} and~\ref{sec:experiments}).
\end{enumerate}

% !TEX root = ../main.tex
\section{Related work}
\label{sec:related}

\subsection{Local geometric estimators and their calibration}
\label{sec:related-estimators}

ID estimators differ in the geometric information they use: projections,
global scaling, and local distances provide distinct starting points~\cite{camastra2016intrinsic,binnie2025survey}. Local principal components
recover affine structure~\cite{fukunaga1971algorithm}, and correlation methods
use fractal scaling~\cite{grassberger1983measuring}. Nearest-neighbor methods
infer dimension from local volume growth. The Levina--Bickel
maximum-likelihood estimator (MLE) uses several neighbor radii through a
local Poisson model~\cite{levina2004maximum}, the two-nearest-neighbor
estimator (TWO-NN) retains only the ratio of the first two neighbor distances~\cite{facco2017estimating}, and Gride extends that ratio to generic neighbor
orders~\cite{denti2022generalized}. Larger orders trade locality for
robustness to short-scale noise, which otherwise inflates the estimate toward
the ambient dimension~\cite{denti2022generalized}.

Angular geometry supplies a complementary source of dimension information, used by the angle-based intrinsic dimensionality (ABID) estimator.
ABID estimates intrinsic dimension
from the second moment of cosine similarities between pairs of normalized
neighbor-displacement directions~\cite{thordsen2022abid}, whereas expected
simplex skewness (ESS) uses simplex geometry~\cite{johnsson2015low}. DANCo models mutual neighbor angles with a von Mises
law whose concentration grows with dimension and combines this angular
information with distance evidence~\cite{ceruti2014danco}. In ABID and DANCo,
angular spread therefore carries the dimension information.

The treatment of finite-sample effects provides a second organizing
distinction. At high generating dimension, neighbors occupy a sizable
fraction of the support, inflating the distance ratio that a local estimator
inverts and producing underestimation~\cite{ceruti2014danco,campadelli2015intrinsic}. One response fits explicit
corrections: the corrected median Farahmand--Szepesv\'ari--Audibert estimator (cmFSA) rescales a
median local estimate~\cite{benko2022manifold}, and GeoMLE combines evidence
across neighborhood sizes~\cite{gomtsyan2019geometry}. Distribution matching
provides another response by placing finite-sample effects on both sides of
the comparison. The Minimum Neighbor Distance KL estimator
(MiND$_{\mathrm{KL}}$) matches distance statistics to uniform-ball
references~\cite{lombardi2011minimum,wang2006nearest}, and DANCo adds angular
KL matching~\cite{ceruti2014danco}. Recent methods correct underestimation~\cite{qiu2022intrinsic,qiu2023underestimation} or learn dimension from neighbor
statistics~\cite{ong2026universal}. Implementations of related estimators are
available in \texttt{scikit-\allowbreak dimension}~\cite{bac2021scikit}, DADApy~\cite{glielmo2022dadapy}, and intRinsic~\cite{denti2023intrinsic}.
Distribution matching is the basis of the calibration studied here.

\subsection{Intrinsic dimension in learned representations}
\label{sec:related-representations}

ID estimates also provide a way to examine geometry across learned
representations. Ansuini et al.~\cite{ansuini2019intrinsic} reported
layer-wise TWO-NN curves that rise, peak, and decline toward the output, with
last-hidden-layer estimates strongly associated with test accuracy. Pope et
al.~\cite{pope2021intrinsic} linked lower dataset ID to lower sample complexity.
Related studies analyze generative and diffusion models~\cite{kamkari2024geometric,stanczuk2024diffusion,choi2025latent} and
transformers~\cite{valeriani2023geometry}, and layer-wise ID profiles also
characterize variational autoencoders, whose curves change shape once the
bottleneck exceeds the data ID~\cite{camboulin2024understanding}.

Interpreting such profiles requires distinguishing the behavior of an
estimator from the dimension of the underlying representation. Schulte and R\"ugamer~\cite{schulte2026rethinking} show that pointwise and Hausdorff dimensions are
nonincreasing through Lipschitz network maps, while rising nearest-neighbor
estimates can reflect distance ratios and representation expansion. This
distinction motivates our use of CNN curves to study representation geometry
and sensitivity to calibration.

% !TEX root = ../main.tex
\section{Method}
\label{sec:method}

Our construction retains DANCo's joint calibration while making each
component available for interpretation and adjustment. Generic-order ratios address neighborhood-relative noise, mean-direction
profiling addresses sample-amplitude heterogeneity, and scaled Bessel
evaluation reaches candidates in the hundreds, within one objective.

Let $\mathcal X=\{x_i\}_{i=1}^{N}\subset\mathbb{R}^{D}$ be sampled near a
$d$-dimensional manifold. Uppercase $X_i$ denotes a random vector and
lowercase $x_i$ its observed value. If $x_{i(j)}$ is the $j$-th nearest
neighbor of $x_i$, define
\begin{equation}
 r_{i,j}=\|x_{i(j)}-x_i\|_2,\qquad
 u_{i,j}=\frac{x_{i(j)}-x_i}{r_{i,j}},\qquad
 \theta_{j\ell}^{(i)}=\arccos\langle u_{i,j},u_{i,\ell}\rangle .
 \label{eq:geometry-definitions}
\end{equation}
Here $N$ is sample size, $D$ is ambient dimension, and $d$ is generating ID
when known. In Eq.~\eqref{eq:geometry-definitions}, the distance $r_{i,j}$
measures local separation, while
$\theta_{j\ell}^{(i)}\in[0,\pi]$ is an angle between neighbor displacements.
The centered norm
$a_i=\|x_i-\bar x\|_2$, with $\bar x=N^{-1}\sum_i x_i$, measures an
observation's displacement from the global mean and is an observable measure
of its amplitude. We use $k$ for the working $k$-nearest-neighbor (kNN)
neighborhood size.

\subsection{Local statistics and DANCo calibration}
\label{sec:method-danco}

Distance calibration will use two published ratio laws: the generic-order
Gride law and DANCo's first-neighbor law. The Gride ratio
$\mu_{i;k_1,k_2}=r_{i,k_2}/r_{i,k_1}$, for neighbor orders $k_1<k_2$,
has the following density under the local Poisson model, where $B(a,b)$ is the beta function:
\begin{equation}
  f(\mu;d,k_1,k_2)
  = \frac{d\,(\mu^{d}-1)^{\,k_2-k_1-1}}{\mu^{\,d(k_2-1)+1}\,B(k_2-k_1,\,k_1)},
  \qquad \mu>1.
  \label{eq:gride-pdf}
\end{equation}
Denti et al.~\cite{denti2022generalized} derive this law; the TWO-NN ratio
of the first two neighbor distances is its
$(k_1,k_2)=(1,2)$ case~\cite{facco2017estimating,denti2022generalized}.
DANCo~\cite{ceruti2014danco} pairs a first-neighbor distance statistic with
angular statistics and calibrates them against samples simulated at each
candidate dimension.

\paragraph{Distance statistic}
For each point, let $\rho_i=r_{i,1}/r_{i,k+1}$. Under the DANCo local model,
this ratio has density
\[
g(\rho;k,d)=kd\rho^{d-1}(1-\rho^{d})^{k-1},\qquad 0<\rho<1.
\]
Maximum likelihood over $\{\rho_i\}$ gives $\dhat_{\mathrm{dist}}$.

\paragraph{Angular statistic}
The $\binom{k}{2}$ pairwise neighbor angles are modeled by
\[
q(\theta;\nu,\tau)=\frac{e^{\tau\cos(\theta-\nu)}}{2\pi I_0(\tau)},
\]
where $I_0$ is the modified Bessel function of the first kind of order zero.
The physical neighbor angles lie in $[0,\pi]$; DANCo embeds them in the circle
and uses this $2\pi$-periodic von Mises model. Mean direction $\nu$ locates the
angles, while concentration $\tau\geq0$ measures their spread and carries the
dimension dependence in the DANCo model.

The circular mean is the direction of the mean resultant vector. For angles
$\alpha_1,\ldots,\alpha_n$, it is computed as
$\atantwo(\sum_r\sin\alpha_r,\sum_r\cos\alpha_r)$~\cite[Sec.~2.3, Eqs.~(2.7)--(2.9)]{fisher1993statistical}. Per-center
$\hat\nu_i$ applies this definition to the $\binom{k}{2}$ angles, while
$\hat\tau_i$ uses Fisher's three-branch maximum-likelihood approximation~\cite[Sec.~3.3.6, Eq.~(3.47)]{fisher1993statistical}. DANCo aggregates centers by
\[
 \hat\nu=\atantwo\!\left(\sum_i\sin\hat\nu_i,
                         \sum_i\cos\hat\nu_i\right),\qquad
 \hat\tau=\frac1N\sum_{i=1}^{N}\hat\tau_i.
\]
When the per-center directions cancel, the circular mean is undefined; such
zero resultants use the distance fallback described in Supplementary
Algorithm S1.

\paragraph{Distribution matching}
DANCo creates one reference for each
candidate $m\in\{1,\ldots,m_{\max}\}$. It simulates $N$ points from the
$m$-dimensional unit hyperball
$\{y\in\mathbb R^m:\|y\|_2\leq1\}$ and records
\[
(\hat d_m^{\mathrm{ref}},\nu_m^{\mathrm{ref}},\tau_m^{\mathrm{ref}}).
\]
The candidate $m$ differs from the unknown $d$ and ambient dimension $D$.
For densities $p$ and $q$, define
\[
 \KL(p\|q)=\int p(t)\log\{p(t)/q(t)\}\,dt.
\]
DANCo's spherical reference model
factorizes the distance and angular laws, so their KL divergences add~\cite{ceruti2014danco}. This joint calibration uses radial volume growth and
directional concentration to support the same candidate dimension. DANCo selects
\begin{equation}
  \dhat \;=\; \argmin_{m}\;
  \KL\!\bigl(g(\cdot;k,\dhat_{\mathrm{dist}})\,\|\,
  g(\cdot;k,\hat d_m^{\mathrm{ref}})\bigr)
  \;+\;
  \KL\!\bigl(q(\cdot;\hat\nu,\hat\tau)\,\|\,
  q(\cdot;\nu_m^{\mathrm{ref}},\tau_m^{\mathrm{ref}})\bigr).
  \label{eq:danco-argmin}
\end{equation}
For $q_a=q(\cdot;\nu_a,\tau_a)$, $a\in\{1,2\}$, the von Mises term is~\cite{vo2008statistical}
\begin{equation}
  \KL(q_1\|q_2) = \log\frac{I_0(\tau_2)}{I_0(\tau_1)}
  + \frac{I_1(\tau_1)}{I_0(\tau_1)}
    \bigl(\tau_1-\tau_2\cos(\nu_2-\nu_1)\bigr),
  \label{eq:klvm}
\end{equation}
where $I_0$ and $I_1$ are modified Bessel functions. Writing $A=I_1/I_0$,
Eq.~\eqref{eq:klvm} separates into a concentration term,
$\log\{I_0(\tau_2)/I_0(\tau_1)\}+A(\tau_1)(\tau_1-\tau_2)$, and the
nonnegative angular-location penalty $A(\tau_1)\tau_2\{1-\cos(\nu_2-\nu_1)\}$.
For a mean-direction gap that is not a multiple of $2\pi$ and positive observed concentration,
the penalty grows with the reference concentration $\tau_2$. When the gap
varies across candidates, its trend depends on both reference parameters; a
rising penalty favors lower candidates even when the concentrations agree.
This separates two influences within angular calibration: agreement in
concentration and the cost of a difference in mean direction.

Optional interpolation around the minimum yields a fractional estimate. We call
Eq.~\eqref{eq:danco-argmin} a composite calibration objective. To reduce
repeated reference-generation cost, FastDANCo replaces
simulation at evaluation time with precomputed smoothing splines over $(N,m)$~\cite[Sec.~4.2]{ceruti2014danco}. Computational complexity and memory bounds
are derived in Supplementary Section S2.2.

\subsection{Componentwise model-based calibration}
\label{sec:method-framework}

To interpret the estimate in Eq.~\eqref{eq:danco-argmin}, we retain its
distance and angular discrepancy curves separately throughout calibration.
Let $\Delta_{\mathrm{dist}}(m)$ and $\Delta_{\mathrm{ang}}(m)$ denote these
discrepancies between the observed statistics and their dimension-$m$
references. Their pointwise sum defines the componentwise estimate:
\begin{equation}
  \dhat = \argmin_{m\in\{1,\ldots,m_{\max}\}} J(m), \qquad
  J(m)=\Delta_{\mathrm{dist}}(m)+\Delta_{\mathrm{ang}}(m).
  \label{eq:composite-calibration}
\end{equation}
The combination retains DANCo's equal weights. We use \emph{composite} for
the summed curve $J$ and \emph{componentwise} for the formulation that keeps
both terms visible.

\begin{figure}[t]
\centering
\includegraphics[width=\linewidth]{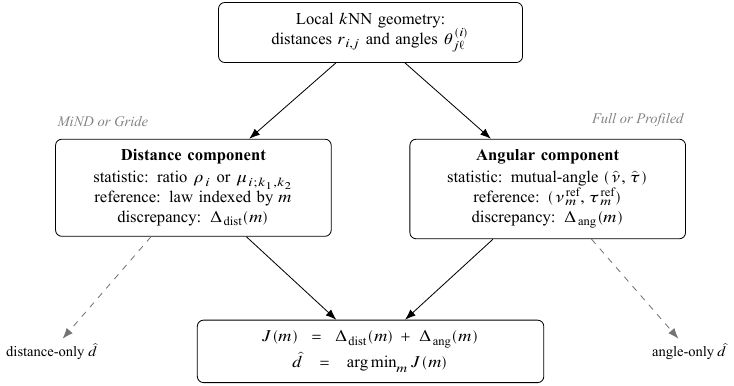}
\caption{Componentwise calibration. Minimum Neighbor Distance (MiND) or the
generalized ratios ID estimator (Gride) supplies the distance discrepancy;
Full or Profiled matching supplies the angular discrepancy. Each curve can be
minimized separately, and their pointwise sum forms the combined objective.
Solid arrows feed the combined objective; dashed branches lead to the
component-only minima. Comparing the three minima shows how each component
affects the estimate.}
\label{fig:framework}
\end{figure}

The componentwise formulation specifies four elements for each component:
observed statistic, dimension-indexed reference, discrepancy, and treatment of
nuisance parameters. The component boxes of Figure~\ref{fig:framework}
show the first three; the labels identify the distance statistic and the
angular nuisance treatment.

MiND--Full is the original DANCo objective. The two choices yield four
variants on identical data, candidate dimensions, and reference construction.
Changing MiND to Gride replaces the distance statistic and reference law;
changing Full to Profiled replaces the angular discrepancy with the profiled form of Eq.~\eqref{eq:profile-general} below, specialized to the angular component in Eq.~\eqref{eq:profile-infimum}. The separate curves
then reveal the influence of each change on the combined minimum.

Nuisance parameters control aspects of the match that need not carry the
target dimension information. For component $c$, let
$\lambda_c\in\Lambda_c$ be such a parameter in the reference. Profiling
compares the data with the best-aligned reference within that family:
\begin{equation}
 \Delta_c^{\mathrm{prof}}(m)
 =\inf_{\lambda_c\in\Lambda_c}\Delta_c(m;\lambda_c).
 \label{eq:profile-general}
\end{equation}
We report both Full and Profiled curves; their difference quantifies the
contribution of mean-direction alignment.

The construction extends beyond these four combinations under a common
admission condition: a statistic enters either slot when its reference law
is indexed by a fitted dimension and carries a computable discrepancy.
The distance component admits analytic examples. For the TWO-NN case
$(k_1,k_2)=(1,2)$ of Eq.~\eqref{eq:gride-pdf}, the factor $k_2-k_1-1$ of the
generic-order divergence derived in Section~\ref{sec:method-gride} vanishes
and the divergence collapses to the closed
form
$\log(\dhat_{\mathrm{dist}}/\hat d_m^{\mathrm{ref}})
+\hat d_m^{\mathrm{ref}}/\dhat_{\mathrm{dist}}-1$. The Levina--Bickel
log-ratio sum $\sum_{j=1}^{k-1}\log(r_{i,k}/r_{i,j})$ is Gamma$(k-1,1/d)$, with shape $k-1$ and scale $1/d$, under the same Poisson-process approximation~\cite{levina2004maximum}. Both examples
therefore enter the calibrated objective without new derivations.

For the angular component, ABID's pairwise-cosine second moment, $\mathbb E[C^2]=1/d$ under its isotropic model~\cite{thordsen2022abid}, can supply a moment-based discrepancy, and simplex statistics such as ESS enter through Monte Carlo reference laws.

%The calibration of one component does not depend on which statistic fills the other slot, so a family of distance statistics shares one angular reference and one angular discrepancy evaluation.

\subsection{Generic-order distance calibration}
\label{sec:method-gride}

DANCo's MiND statistic, $\rho_i=r_{i,1}/r_{i,k+1}$, anchors the distance
comparison at the first neighbor. To reduce sensitivity to short-scale
noise, we substitute the Gride ratio
$\mu_{i;k_1,k_2}=r_{i,k_2}/r_{i,k_1}$ from Eq.~\eqref{eq:gride-pdf}, using
$k_1=\lceil k/2\rceil$ and $k_2=2k_1$. The largest neighbor orders are
comparable: MiND uses order $k+1$, whereas Gride uses order $k$ when $k$ is
even. Gride supplies the generic-order ratio law and likelihood estimator~\cite{denti2022generalized}. Integrating this statistic into
Eq.~\eqref{eq:composite-calibration} requires a divergence between its fitted
data law and each finite-sample reference law, which we derive below.

We fit the observed dimension $\dhat_{\mathrm{dist}}$ with the published Gride
maximum-likelihood procedure~\cite{denti2022generalized}. Applying the same
procedure to the reference sample simulated at candidate dimension $m$ gives
$\hat d_m^{\mathrm{ref}}$.

Let the data and dimension-$m$ reference laws be
$f=f(\cdot;\dhat_{\mathrm{dist}},k_1,k_2)$ and
$f_m=f(\cdot;\hat d_m^{\mathrm{ref}},k_1,k_2)$, and define
$\gamma_m=\hat d_m^{\mathrm{ref}}/\dhat_{\mathrm{dist}}>0$. Their KL divergence is
\begin{equation}
\begin{split}
\KL(f\|f_m)
={}& \log\frac{\dhat_{\mathrm{dist}}}{\hat d_m^{\mathrm{ref}}}
+ (k_2-1)\Bigl(\frac{\hat d_m^{\mathrm{ref}}}{\dhat_{\mathrm{dist}}}-1\Bigr)
  \bigl(\Psi(k_2)-\Psi(k_1)\bigr)\\
&+ (k_2-k_1-1)\Bigl[\Psi(k_2-k_1)-\Psi(k_1)
-\mathcal{I}(\gamma_m)\Bigr].
\end{split}
\label{eq:klgride}
\end{equation}
Here $\Psi$ is the digamma function. For the integer neighbor orders used throughout, the beta-prime expectation $\mathcal I(\gamma_m)$ reduces to a finite sum of $k_2-k_1$ digamma terms, so the divergence is available in closed form. Supplementary Section S1.1 derives the transformation, the closed form, and the check $\KL(f\|f)=0$.
Substituting the ratio changes only the distance slot of
Eq.~\eqref{eq:composite-calibration}; the angular statistic and the
equal-weight combination are unchanged, so any accuracy difference from
MiND--Full isolates the distance statistic.

\subsection{Angular concentration and profiled divergence}
\label{sec:method-nustrip}

Angular calibration compares both concentration $\tau$ and mean direction
$\nu$, but these parameters have different geometric roles. Distinguishing
them explains when location alignment can preserve useful concentration
matching. We develop this distinction in two sampling regimes, with proofs
in Supplementary Section S1.2.

\begin{proposition}[The mean direction is dimension-free under symmetry]
\label{prop:nu-symmetric}
Let $u,v$ be independent uniform directions in $\mathbb{R}^{d}$ and
$\theta$ the angle between them. Then $\theta$ has density proportional to
$\sin^{d-2}\theta$ on $[0,\pi]$~\cite[Lemmas~11--12, Eqs.~(12)--(13)]{cai2013distributions}, symmetric about
$\pi/2$; in particular $\mathbb{E}[\theta]=\pi/2$ exactly for every $d\ge2$.
\end{proposition}

Proposition~\ref{prop:nu-symmetric} applies to a manifold neighborhood under
the local-symmetry condition of Supplementary Section S1.2: at an interior point, with $d$
and $k$ fixed while $N\to\infty$, the normalized kNN displacements converge to
independent uniform directions on the unit sphere of the tangent space. Taking
the sample limit before $d$ grows then gives angles with mean $\pi/2$ and
variance of order $1/d$. A data--reference gap in the finite-sample
$\hat\nu$ then diagnoses a mismatch with the calibrated reference geometry.
Concentration retains the dimension information posited by the DANCo model,
$\tau\approx d$~\cite{ceruti2014danco}.

The second regime fixes $N$ and $k$ while $d\to\infty$, so neighborhoods can
span nonlocal geometry. The following lemma states the thin-shell and
near-orthogonality limits used throughout this regime. We write
$\xrightarrow{p}$ for convergence in probability.

\begin{lemma}[Thin-shell concentration and near-orthogonality]
\label{lem:thin-shell}
For each $s$ in a fixed finite index set, let
$Z_{s,1},Z_{s,2},\ldots$ be an independent and identically distributed
(i.i.d.) sequence with mean zero, variance one, and finite fourth moment.
Assume the sequences are mutually independent, and let
$Z_s^{(d)}=(Z_{s,1},\ldots,Z_{s,d})\in\mathbb R^{d}$ be the vector of its
first $d$ coordinates. Then,
simultaneously over the fixed collection,
\[
 \frac{\|Z_s^{(d)}\|^2}{d}\xrightarrow{p}1,
 \qquad
 \frac{\langle Z_s^{(d)},Z_t^{(d)}\rangle}{d}\xrightarrow{p}0
 \quad(s\ne t).
\]
\end{lemma}

Lemma~\ref{lem:thin-shell} places observations on an asymptotically thin shell
with nearly orthogonal directions. These limits are the elementary
coordinatewise case of standard high-dimension, low-sample-size geometry~\cite{hall2005geometric}; Supplementary Section S1.2 gives the proof.
The estimator uses angles between displacements from a shared center. The
next proposition translates the concentration limits into that geometry,
including unequal shell radii.

\begin{proposition}[Displacement-angle limit under norm and inner-product concentration]
\label{prop:shell-limit}
For each $d$, let $X_s^{(d)}\in\mathbb R^d$ and let
$\theta_{j\ell}^{(i,d)}$ be the angle between
$X_j^{(d)}-X_i^{(d)}$ and $X_\ell^{(d)}-X_i^{(d)}$. Suppose there are
nonnegative constants $R_s$ such that, simultaneously for fixed $i,j,\ell$,
$\|X_s^{(d)}\|^2/d\xrightarrow{p}R_s^2$ for $s\in\{i,j,\ell\}$ and
$\langle X_s^{(d)},X_t^{(d)}\rangle/d\xrightarrow{p}0$ for $s\ne t$, with
$(R_i^2+R_j^2)(R_i^2+R_\ell^2)>0$. Then
\begin{equation}
 \cos\theta_{j\ell}^{(i,d)}\xrightarrow{p}
 \frac{R_i^2}{\sqrt{(R_i^2+R_j^2)(R_i^2+R_\ell^2)}}.
 \label{eq:shell-angle-limit}
\end{equation}
With fixed $N$, the convergence is simultaneous over all triples whenever the
assumptions hold simultaneously.
\end{proposition}

\begin{figure}[t]
\centering
\includegraphics[width=\linewidth]{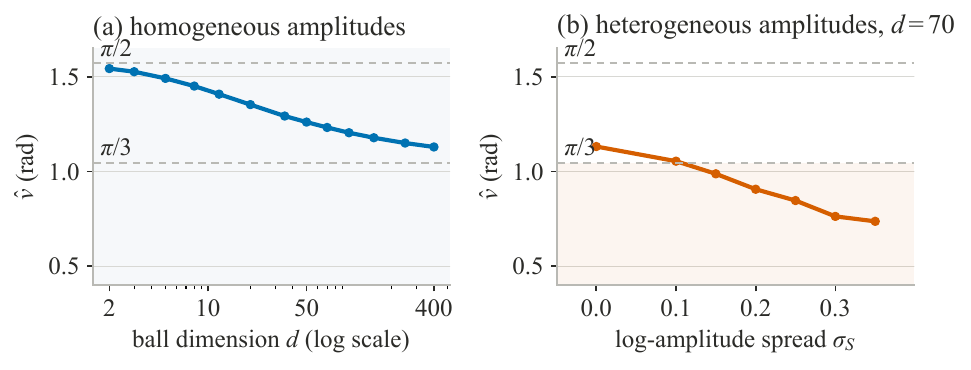}
\caption{Finite-sample checks of the two angular regimes ($N{=}2500$,
$k{=}10$). Dashed lines mark the local-interior mean
$\pi/2$ and homogeneous-shell limit $\pi/3$. (a) Uniform $d$-balls as $d$
varies at fixed sample size. (b) Gaussian scale mixtures at $d=70$ as the
log-amplitude spread $\sigma_S$ varies over the grid of the Section~\ref{sec:exp-gsm} sweep, the range anchored by the image classes.}
\label{fig:nu-mechanism}
\end{figure}

On a homogeneous thin shell, $R_i=R_j=R_\ell$, so
$\theta\xrightarrow{p}\pi/3$. This connects the classical concentration of
high-dimensional samples near a shell with the displacement-angle geometry
used by DANCo. Supplementary Corollaries S1.5--S1.7 specialize the result to
i.i.d. coordinates, uniform hyperballs, and Gaussian scale mixtures. Figure~\ref{fig:nu-mechanism} illustrates the transition from the local-interior regime toward this limit.

The effect of unequal amplitudes can be isolated without changing generating
dimension. Define the Gaussian scale mixture (GSM)
\[
 X_s^{(d)}=S_sZ_s^{(d)},\qquad
 Z_s^{(d)}\overset{\mathrm{i.i.d.}}\sim\mathcal N(0,\mathbf I_d),\qquad
 \log S_s\overset{\mathrm{i.i.d.}}\sim\mathcal N(0,\sigma_S^2).
\]
Here $S_s>0$ is the multiplicative amplitude of one observation,
$\mathbf I_d$ is the
$d\times d$ identity matrix, and $\sigma_S$ is the standard deviation of log
amplitude. The amplitude and Gaussian sequences are mutually independent.

Unequal amplitudes preserve ID but affect neighbor selection. In a Gaussian
scale mixture, the pairwise distance obeys
\[
 d^{-1}\|X_j^{(d)}-X_i^{(d)}\|_2^2\xrightarrow{p}S_i^2+S_j^2.
\]
For a fixed center, $S_i^2$ is common to all limiting distances, so
lower-amplitude observations tend to be closer and are preferentially selected
as neighbors.

Proposition~\ref{prop:shell-limit} gives the angular
consequence. When both selected-neighbor amplitudes are below the center
amplitude, the numerator in Eq.~\eqref{eq:shell-angle-limit} remains $S_i^2$,
while each denominator factor is below $2S_i^2$. The limiting cosine therefore
exceeds $1/2$, and the angle falls below $\pi/3$.

Figure~\ref{fig:nu-mechanism} tests the finite-sample transition between the
local-interior and shell regimes and then isolates the location shift
from sample-amplitude heterogeneity. The sampled mean direction leaves $\pi/2$
and approaches $\pi/3$ as $d$ grows, and falls below $\pi/3$ as $\sigma_S$
grows. At the center level, a larger centered norm goes with a smaller per-center mean direction: in Eq.~\eqref{eq:shell-angle-limit} with $R_s=S_s$, raising the center amplitude relative to its selected neighbors raises the limiting cosine and lowers the angle. Supplementary Figure S1 shows this monotone decrease across
centered-norm deciles.

Because physical neighbor angles occupy $[0,\pi]$ whereas DANCo uses a circular
von Mises approximation, Supplementary Section S1.4 contrasts the exact locally
symmetric angle density with the movable-location von Mises family and then
assesses separately fitted pooled von Mises laws (Figure S2). Their mass outside $[0,\pi]$ is at most $5.0\times10^{-8}$, and the fitted laws pass descriptive checks of fit.

The two regimes distinguish a dimension-bearing concentration from a location
that also responds to geometry and amplitude. This motivates profiling
angular location while retaining concentration matching, recognizing
that concentration may itself be affected by the changed geometry.
Following Eq.~\eqref{eq:profile-general}, let $\delta\in[0,2\pi)$ rotate the
reference mean direction. Comparing the von Mises laws modulo this rotation gives
\begin{equation}
\Delta_{\mathrm{ang}}^{\mathrm{prof}}(m)
=\inf_{\delta}\,
\KL\bigl(q(\cdot;\hat\nu,\hat\tau)\,\|\,
q(\cdot;\nu_m^{\mathrm{ref}}+\delta,\tau_m^{\mathrm{ref}})\bigr).
\label{eq:profile-infimum}
\end{equation}
Aligning their mean directions gives
\begin{equation}
\Delta_{\mathrm{ang}}^{\mathrm{prof}}(m)
= \log\frac{I_0(\tau_m^{\mathrm{ref}})}{I_0(\hat\tau)}
+ A(\hat\tau)\,(\hat\tau-\tau_m^{\mathrm{ref}}).
\label{eq:klstrip}
\end{equation}
Supplementary Section S1.3 derives the profiled infimum in
Eq.~\eqref{eq:profile-infimum} and its closed form in Eq.~\eqref{eq:klstrip}.
The profiled angular discrepancy therefore depends on the observed angular
statistics only through $\hat\tau$; it equals the concentration term of
Eq.~\eqref{eq:klvm}, so profiling removes the angular-location penalty exactly.
When concentration remains informative, this removes the influence of
location mismatch while preserving concentration evidence. Full matching
retains additional information when data and references share the same
finite-sample geometry.

\subsection{Numerical evaluation}
\label{sec:method-impl}

The four variants share the neighbor geometry, reference construction, and
curve-minimization procedure. Stable angular evaluation carries this
comparison into the candidate range of learned representations.
At reference sample size $500$, ordinary Bessel evaluation first overflows at candidate $m=278$, and the boundary is similar across the precomputed sample sizes $450$--$700$. Exponentially
scaled Bessel evaluation keeps both angular objectives finite through the
CNN study's bound $m_{\max}=400$. Supplementary Section S2.3 defines the
scaled identities and reports the observed overflow boundary.

To avoid regenerating one reference per candidate for every CNN layer, we
precompute FastDANCo-style smoothing-spline reference surfaces. One
offline procedure builds a surface for each distance statistic on a shared
simulation grid (Supplementary Section S3.7). Supplementary
Algorithm S1 (Section S2.1) states the complete estimator, including our chosen
low-dimension rule: a preliminary distance fit at most five is returned directly,
without reference matching. Section S2.4
maps each computational measurement to its timed steps.
Supplementary Figures S3--S4 show the overflow boundary and workload-aligned
central-processing-unit (CPU) and graphics-processing-unit (GPU) scaling. At
sample size 500, ambient dimension 400, and ten
neighbors, complete evaluation through candidate 400 with precomputed references
finishes in well under one second on both devices.

% !TEX root = ../main.tex
\section{Experiments}
\label{sec:experiments}

The experiments examine when the two geometric components support a common
estimate and how targeted changes improve calibration. Clean manifolds test
accuracy and component agreement; neighborhood-relative noise tests the
distance statistic; and a known-$d$ GSM tests profiling under
sample-amplitude heterogeneity. Natural images and CNN representations, with candidates through $400$, then show what
the component statistics reveal when generating dimension is unknown.

\subsection{Experimental design and metrics}
\label{sec:exp-setup}

\paragraph{Estimators and datasets}
We combine MiND or Gride distance calibration with Full or Profiled angular
calibration. Distance-only and angle-only objectives isolate the contribution
of each component. Calibrated objectives use $k=10$ except where a sweep varies it; Gride uses neighbor
orders $(k_1,k_2)=(5,10)$. The primary benchmark uses the 24 manifolds returned
by \texttt{scikit-dimension}'s \texttt{BenchmarkManifolds} generator, whose
suite extends the benchmark introduced by Campadelli et al.~\cite{campadelli2015intrinsic,bac2021scikit}.

The nine reference estimators are the Levina--Bickel maximum-likelihood
estimator (MLE)~\cite{levina2004maximum}, TWO-NN, computed throughout with
its original discarded-tail fit~\cite{facco2017estimating},
local principal component analysis (lPCA)~\cite{fukunaga1971algorithm},
expected simplex skewness (ESS)~\cite{johnsson2015low},
manifold-adaptive dimension estimation (MADA)~\cite{farahmand2007manifold}, the
tight local estimator (TLE)~\cite{amsaleg2019tight}, correlation dimension
(CorrInt)~\cite{grassberger1983measuring}, MiND maximum likelihood
(MiND--ML)~\cite{lombardi2011minimum}, and Fisher separability
(FisherS)~\cite{albergante2019estimating}.

For the primary benchmark over collection $\mathcal B$,
let $d_M$ denote the generating ID of
manifold $M\in\mathcal B$ and let $\hat d_M$ denote its estimate averaged over
replicates. We report
mean percentage error (MPE)~\cite{lombardi2011minimum},
\[
 \mathrm{MPE}=\frac{100}{|\mathcal B|}\sum_{M\in\mathcal B}
 \frac{|\hat d_M-d_M|}{d_M}.
\]
The sensitivity, component, noise, and GSM comparisons compute MPE within
each replicate before summarizing across replicates. Supplementary Section S3 defines
the datasets, comparison units, and aggregation rules for every experiment,
and Table S1 reports the sample-size-specific angular reference ranges.

\paragraph{Controlled data models}
The amplitude experiment uses the GSM defined in
Section~\ref{sec:method-nustrip}. We set $d=70$, embed each sample in
$D=100$ so that the candidate search covers $1$--$100$, and evaluate
$\sigma_S\in\{0,0.1,0.15,0.2,0.25,0.3,0.35\}$ over 30 replicates, a range that
brackets the centered-norm variation of the 17 image classes surveyed in
Supplementary Table S9; configurations share data and reference replicates
(Supplementary Section S3.1).

Profiling tests whether removing the angular-location penalty improves the
combined estimate. A second
control divides each observation by its known simulated amplitude,
$X_i/S_i=Z_i$. Within each replicate, this operation holds $Z_i$, the
generating dimension, and the reference replicate fixed. Multiplicative
amplitude is the only generating factor removed, so recovery of Full
calibration identifies its effect within the GSM.

For additive noise, let $X_i^{(0)}\in\mathbb R^D$ be the clean observation and
let $r_{i,10}^{(0)}$ be its Euclidean distance to the tenth nearest neighbor in
the clean sample. Conditional on that sample, we observe
\[
 X_i^{\mathrm{obs}}=X_i^{(0)}+\varepsilon_i,
 \qquad \varepsilon_i\overset{\mathrm{i.i.d.}}\sim
 \mathcal N(0,\sigma_\varepsilon^2 \mathbf I_D),
\]
where $\mathbf I_D$ is the $D\times D$ identity matrix. We set the coordinate scale by
\begin{equation}
 \sigma_\varepsilon
 =\eta\,\frac{\operatorname{median}_i r_{i,10}^{(0)}}{\sqrt{2D}}.
 \label{eq:relative-noise}
\end{equation}
The dimensionless parameter $\eta$ measures noise relative to local neighbor
spacing. Conditional on the clean sample, two independent noise vectors satisfy
\[
 \mathbb E\|\varepsilon_i-\varepsilon_j\|_2^2
 =2D\sigma_\varepsilon^2
 =\eta^2\{\operatorname{median}_i r_{i,10}^{(0)}\}^2.
\]
Thus the median spacing adapts the perturbation to the dataset's global scale,
and $\sqrt{2D}$ removes the ambient-dimensional growth of the root-mean-square
noise displacement. At $\eta=1$, that displacement equals the median clean
tenth-neighbor distance. We measure error against the pre-noise generating
dimension (Supplementary Section S3.4).

The image experiments assess calibration sensitivity through two descriptive
normalizations. Centered radial
normalization first sets $c_i=x_i-\bar x$ within a class and maps every
nonzero $c_i$ to $c_i/\|c_i\|_2$. Per-image contrast normalization maps
coordinate $j$ to $x'_{ij}=(x_{ij}-\bar x_i)/s_i$, where $\bar x_i$ and $s_i$
are that image's coordinate mean and standard deviation. Both transformations
reduce sample-amplitude variation and support the reference-mismatch
interpretation of Section~\ref{sec:exp-diagnostics}
(Supplementary Section S3.6).

\subsection{Clean-manifold accuracy}
\label{sec:exp-benchmark}

The clean benchmark establishes the accuracy of the joint calibration on
which the componentwise formulation builds. Under the shared
twenty-replicate design, MiND--Full has the lowest MPE in
Table~\ref{tab:baselines}: $6.33\%$, followed by TWO-NN at $11.12\%$ and
MiND--ML at $13.58\%$. This agrees with earlier comparisons placing DANCo
among the strongest ID estimators~\cite{campadelli2015intrinsic,denti2022generalized}. MiND--Full's error rate,
the fraction of individual estimates with more than $10\%$ relative error, is
$0.138$. Accuracy, cost, and reliability distinguish the alternatives: ESS reaches
$20.20\%$ MPE at a median $101.61$ seconds per dataset, and FisherS returns 19
failed cells. Supplementary Section S4.1, Table S2, and Figure S5 report the
per-manifold estimates and signed-error curves. These distinguish MiND--Full's
high-dimensional accuracy from its remaining geometry-specific errors.
Supplementary Section S4.3 and Table S4 record the neighborhood and
sample-size sweeps specified in Section S3.3. There the two MiND objectives remain the top pair of
the five compared methods across neighborhood sizes, and every calibrated
configuration improves with sample size.

\begin{table}[t]
\centering
\caption{Ten intrinsic-dimension estimators on 24 manifolds with 20 data
replicates per estimator. MPE averages each manifold's estimate over its
successful replicates and then averages the relative errors over the 24
manifolds; failures count as errors in the error rate. The last column reports
the median time in seconds to estimate one dataset on an Intel i7-10700 CPU,
and MiND--Full excludes the shared reference construction.}
\label{tab:baselines}
\vspace{4pt}
\normalsize
\begin{tabular}{l rrrr}
\toprule
Estimator & MPE (\%) & $>10\%$ error & failed & median time (s) \\
\midrule
MiND--Full (DANCo) & 6.33 & 0.138 & 0 & 0.03 \\
TWO-NN & 11.12 & 0.442 & 0 & 0.01 \\
MiND--ML & 13.58 & 0.537 & 0 & 0.06 \\
MLE & 18.84 & 0.596 & 0 & 0.05 \\
ESS & 20.20 & 0.417 & 0 & 101.61 \\
TLE & 20.45 & 0.667 & 0 & 0.51 \\
MADA & 28.71 & 0.594 & 0 & 0.42 \\
FisherS & 37.97 & 0.613 & 19 & 0.44 \\
CorrInt & 38.19 & 0.660 & 0 & 0.08 \\
lPCA & 49.81 & 0.625 & 0 & $<0.01$ \\
\bottomrule
\end{tabular}

\end{table}

A separate comparison isolates the contribution of combining distance and
angle on matched clean geometry. It retains the 14 manifolds on which all
eight objectives are defined; the preliminary-distance cutoff excludes the
other ten (Supplementary Section S3.2). With five shared
reference replicates for every objective, MiND--Full has the lowest
observed mean MPE at $8.6\pm0.7\%$, below MiND distance-only at
$10.1\pm0.2\%$ and Full angle-only at $15.9\pm0.7\%$. MiND--Profiled also
improves on both corresponding components: its MPE is $9.7\pm0.2\%$, below
MiND distance-only at $10.1\pm0.2\%$ and Profiled angle-only at
$12.2\pm0.3\%$. Both combined objectives therefore have lower observed mean
error than their corresponding component-only objectives, with MiND--Full
lowest on this comparison set.

The combined signed errors in Table S3 are each closer to the distance-only values than to the angle-only values. Equal objective weights need not imply equal local influence: a local quadratic analysis of the sampled reference curves supports stronger distance influence for the Profiled objectives. Supplementary Section S4.2 and Table S3 give the derivation, the numerical diagnostics, and all eight objectives.

\subsection{Robustness to neighborhood-relative noise}
\label{sec:exp-noise}

\begin{figure}[t]
\centering
\includegraphics[width=0.65\linewidth]{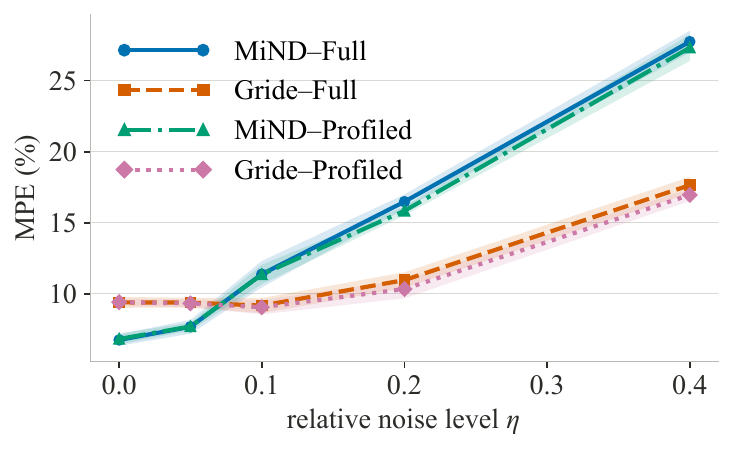}
\caption{MPE over 24 manifolds under Gaussian noise scaled by
Eq.~\eqref{eq:relative-noise} (ten data replicates; bands show one
standard deviation). MiND--Full has lower clean-data error; Gride--Full has
lower error for $\eta\geq0.1$. Color, line style, and marker jointly identify
each method.}
\label{fig:noise-curves}
\end{figure}

The noise experiment tests the practical consequence of replacing the first
neighbor with larger orders. Figure~\ref{fig:noise-curves} compares the
combined estimates as noise increases relative to neighborhood spacing.

Figure~\ref{fig:noise-curves} identifies the conditions favoring the larger-order
ratio. MiND--Full leads on clean data ($6.75\%$ versus $9.39\%$ MPE under
this ten-replicate design). At $\eta=0.1$, Gride--Full reaches $9.16\%$, compared with
$11.38\%$ for MiND--Full; at $\eta=0.4$, the corresponding errors are $17.64\%$
and $27.74\%$. The larger-order ratio improves accuracy for $\eta\geq0.1$
within the evaluated range. Supplementary Section S4.4 and Table S5 report
the full comparison, in which both Gride-based combined objectives degrade more
slowly with $\eta$ than their MiND counterparts.

\subsection{Angular-reference mismatch}
\label{sec:exp-images}

We examine the angular component first in a setting with known generating
dimension, where the GSM isolates amplitude variation, and then on natural
images. This separates a controlled accuracy test from the descriptive
question of how reference mismatch appears in observed data.

\subsubsection{Known-dimension Gaussian scale mixture}
\label{sec:exp-gsm}

\begin{figure}[t]
\centering
\includegraphics[width=\linewidth]{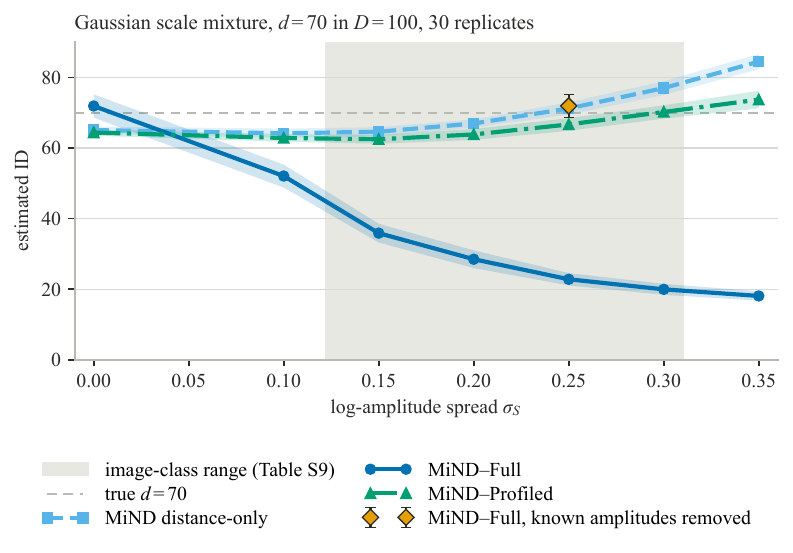}
\caption{Gaussian scale-mixture sweep at $d=70$ embedded in $D=100$ over 30
data replicates. The log-amplitude standard deviation is $\sigma_S$;
$S_i>0$ is the simulated amplitude in $X_i=S_iZ_i$, and $Z_i$ is a
standard Gaussian vector before embedding. Curves show the mean estimated ID with bands of $\pm$ one
standard deviation; the shaded band approximately maps the centered-norm
variation of the 17 surveyed image classes (Supplementary Table S9) onto the
log-amplitude spread; the diamond is MiND--Full after the known-amplitude
control $X_i/S_i=Z_i$ at $\sigma_S=0.25$; the dashed line marks the generating
dimension.}
\label{fig:fusion-gsm}
\end{figure}

The GSM connects the accuracy benefit of angular profiling to its removal
of the location penalty
(Figure~\ref{fig:fusion-gsm}). At $\sigma_S=0.25$, the focal level inside the
image-class range, MiND--Full estimates $22.80\pm1.76$ and has
$67.43\%$ MPE, whereas MiND--Profiled estimates $66.67\pm1.81$ ($4.76\%$
MPE) and MiND distance-only $71.20\pm1.75$ ($2.48\%$ MPE) against the generating
dimension 70. Profiling leaves the distance objective unchanged and
removes the angular-location penalty, returning the combined estimate to the
distance evidence.

The effect follows the penalty in
Section~\ref{sec:method-danco}: the observed mean direction, $0.832\pm0.014$,
has no candidate reference mean direction below it. Over five focal-level replicates, the angular-location penalty rises across nearly all candidates (Supplementary Section S4.5). This penalty shifts the joint minimum to $22.80$;
profiling removes it exactly.

The sweep establishes the extent of this benefit. Across the seven tested
levels, MiND--Full falls monotonically from $71.90$ at
$\sigma_S=0$ to $18.08$ at $\sigma_S=0.35$, while MiND--Profiled stays near the generating dimension with means between $62.47$ and $73.70$. Profiling therefore
brings the combined estimate closer to the generating dimension at every
nonzero amplitude level (Supplementary Sections S3.5 and S4.5 and
Tables S6--S7).

The amplitude-removal control at $\sigma_S=0.25$ moves the mean direction
from $0.832\pm0.014$ to $1.137\pm0.003$ radians and MiND--Full from
$22.80\pm1.76$ to $71.90\pm3.30$, recovering the amplitude-free $\sigma_S=0$
sample exactly; within the GSM, Full underestimation is thus attributable to the amplitudes alone.

\subsubsection{Natural images and normalization}
\label{sec:exp-diagnostics}

Natural images test whether the angular signature identified in the GSM is
also visible in data with uncontrolled amplitude structure. Gaussian scale
mixtures model multiscale image coefficients~\cite{wainwright1999scale},
motivating images as an observational counterpart to the controlled study.

Table~\ref{tab:image-stats-main} reports all four variants on the six raw
image classes. MNIST lies inside its sample-size-specific reference
range, its $\hat\nu$ exceeds one, and the four variants agree: MiND--Full and
Gride--Full agree to the reported precision, and the reported means of the two
Profiled objectives agree within $0.6$ dimensions ($20.20$ versus $20.00$ for
digit 3 and $14.80$ versus $14.20$ for digit 7). CIFAR-10 and ImageNet
have mean directions below their reference ranges and below one, with larger Full--Profiled separations, and the two Full estimates separate ($18.78$
versus $15.80$ on CIFAR-10 bird). This pattern is consistent with angular-location mismatch, which also makes the Full estimate depend on the distance statistic. Gride--Profiled combines the generic-order distance
statistic with angular-location profiling and returns the lowest profiled
estimate on these four CIFAR-10 and ImageNet classes.

\begin{table}[t]
\centering
\caption{Estimated ID and angular statistics on six raw image classes.
Calibrated estimates are means over five reference replicates; $\hat\nu$ is
the observed mean direction in radians and $\hat\tau$ is the dimensionless
observed von Mises concentration. Reference-replicate standard
deviations reach $2.1$ for the Profiled objectives and $4.6$ for the Full
objectives.}
\label{tab:image-stats-main}
\vspace{4pt}
\normalsize
\setlength{\tabcolsep}{4pt}
\begin{tabular}{l rrrrr rr}
\toprule
 & \multicolumn{2}{c}{Full} & \multicolumn{2}{c}{Profiled} & & & \\
\cmidrule(lr){2-3}\cmidrule(lr){4-5}
Dataset & MiND & Gride & MiND & Gride & TWO-NN & $\hat\nu$ & $\hat\tau$ \\
\midrule
MNIST digit 3 & 20.39 & 20.39 & 20.20 & 20.00 & 14.88 & 1.166 & 41.3 \\
MNIST digit 7 & 14.40 & 14.40 & 14.80 & 14.20 & 12.19 & 1.157 & 27.7 \\
CIFAR-10 bird & 18.78 & 15.80 & 34.00 & 33.20 & 25.55 & 0.898 & 63.2 \\
CIFAR-10 cat & 21.21 & 17.60 & 32.80 & 32.20 & 26.11 & 0.946 & 56.8 \\
ImageNet koala & 24.86 & 20.71 & 45.00 & 38.80 & 31.32 & 0.934 & 90.1 \\
ImageNet butterfly & 28.07 & 22.20 & 58.80 & 54.60 & 36.11 & 0.915 & 134.6 \\
\bottomrule
\end{tabular}

\end{table}

The Profiled estimates, larger than the raw MiND--Full and TWO-NN values, lie within the range reported for image data by other ID approaches. Earlier pixel-space studies report estimates on the tens-of-dimensions scale: Pope et al.~\cite{pope2021intrinsic} give dataset-level MLE ranges of $7$--$13$ for MNIST, $13$--$26$ for CIFAR-10, and $26$--$43$ for ImageNet across neighborhood sizes, and Ansuini et al.~\cite{ansuini2019intrinsic} report class-level ImageNet estimates on that scale, although the estimators and sampling units differ from ours. More recent diffusion-based estimators give substantially larger values: for MNIST, Stanczuk et al.~\cite{stanczuk2024diffusion} report $66$--$152$ per digit, compared with $13.3$--$14.1$ from dataset-level MLE at the neighborhood sizes considered in that study, with average diffusion-based estimates near $130$ in related work~\cite{kamkari2024geometric}. Estimates above classical local-MLE values are therefore not by themselves anomalous.

The normalization comparison shows how angular-location calibration affects
the reported estimates. For ImageNet koala, contrast normalization moves
$\hat\nu$ from $0.934$ to $1.080$ and the MiND--Full/MiND--Profiled estimates
from $24.9/45.0$ to $72.2/70.0$; the shift narrows the Full--Profiled gap from
$20.1$ to $2.2$, consistent with angular-location mismatch contributing to the
raw-image discrepancy. On the same raw class, Gride--Profiled gives $38.80$,
which is $6.2$ dimensions below MiND--Profiled, showing the additional
effect of the distance-statistic choice within the profiled objective.
Supplementary Section S3.6 and Table S1
report the reference ranges; Supplementary Section S4.6, Tables S8--S9, and
Figures S6--S7 report the normalization controls, component curves, per-center
relation, and additional classes. Normalization moves angular
location toward the reference range and narrows the Full--Profiled separation
in each CIFAR-10 and ImageNet class of Table S8.

\subsection{Layer-wise profiles in convolutional neural networks}
\label{sec:exp-cnn}

Layer-wise analysis extends the calibration question from pixel-space images
to learned representations. Prior work traces ID through network depth~\cite{ansuini2019intrinsic}; here the candidate range reaches the hundreds.
Gride--Profiled suits this setting: Section~\ref{sec:exp-noise} found the generic-order ratio more robust to noise, and the layer mean directions reported below fall under the reference mean-direction range, the regime that profiling addresses.
The CNN study evaluates AlexNet~\cite{krizhevsky2012imagenet},
VGG-16~\cite{simonyan2015very}, and ResNet-18/34~\cite{he2016deep} on
subsamples of 500 images. At each checkpoint, Gride--Profiled, TWO-NN, and MLE share the activation matrix after exact-duplicate rows are removed; Gride--Profiled evaluates the precomputed Gride reference surface of
Section~\ref{sec:method-impl} at the resulting sample size. Curves are plotted
against relative depth, which places every network on a common input-to-output
interval (Supplementary Section S3.7).

\begin{figure}[t]
\centering
\includegraphics[width=0.90\linewidth]{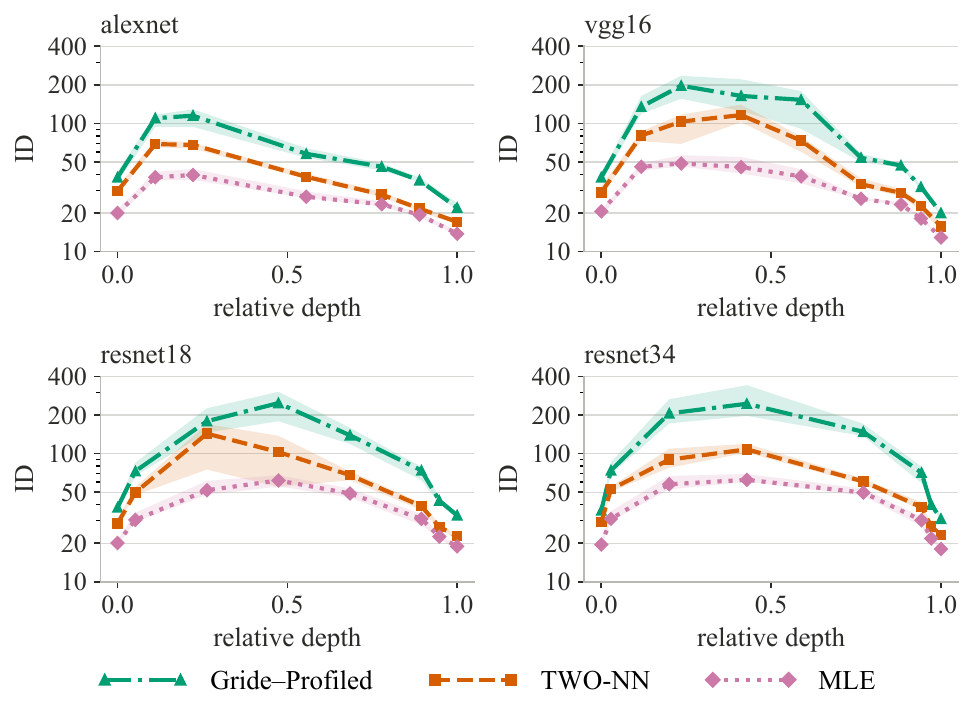}
\caption{Layer-wise ID profiles on identical activations in four pretrained
CNNs. Gride--Profiled uses $k=10$ and precomputed smoothing-spline references
through candidate 400. Lines show the median over seven ImageNet categories
and three 500-image subsamples; bands show their interquartile range. Color,
line style, and marker jointly identify each estimator. The ID axis is
logarithmic.}
\label{fig:cnn-hump}
\end{figure}

The three estimators share a rise-and-fall profile: Gride--Profiled, TWO-NN,
and MLE rise through intermediate layers and decline toward the output
(Figure~\ref{fig:cnn-hump}). Among the sampled checkpoints, the Gride--Profiled
peak falls at the second max-pooling checkpoint of AlexNet and VGG-16, at
$22.2$--$23.5\%$ relative depth, and the second residual stage of ResNet-18 and
ResNet-34, at $42.9$--$47.4\%$. These checkpoints place the peak before the
later pooling or residual stages. This early peak is consistent with near-one neighbor-distance ratios inflating the estimate in learned representations~\cite{schulte2026rethinking}. Further downsampling coincides with
descending estimates toward the output, the same early-peak-then-compression course that
Ansuini et al.\ report for TWO-NN~\cite{ansuini2019intrinsic}. The
shared rise-and-fall shape recurs across categories, subsamples, and the three estimators, which
supports interpreting profile shape and calibration sensitivity (Supplementary Section S4.7).

In every architecture, the largest median paired Full--Profiled separation
sits at the Profiled peak, where $\hat\nu$ lies between $0.80$ and $0.92$:
Gride--Full reports $17$--$37$ while Gride--Profiled reports $115$--$248$, a median paired difference of $-100$ to $-214$ dimensions (Table S10). Layer medians of $\hat\nu$ span
$0.780$--$1.106$ and stay below the CNN reference surface's mean-direction
range (Supplementary Section S3.7). They reach $0.78$--$0.87$ at the first pooling checkpoint of each network and rise to $1.04$--$1.11$ toward the output,
where the two objectives converge (Table S10).
Supplementary Section S4.7 and Table S10 report the peak layer, $\hat\nu$,
Gride--Full, Gride--Profiled, and their difference for each architecture.

% !TEX root = ../main.tex
\section{Discussion and conclusion}
\label{sec:discussion}

On clean benchmarks, joint calibration is among the state of the art (lowest MPE of ten estimators, Table~\ref{tab:baselines}), and MiND--Full ($8.6\pm0.7\%$ MPE) improves on both components on 14 manifolds. Componentwise calibration keeps this strength and addresses each practical condition: neighborhood-relative noise by generic-order ratios, sample-amplitude heterogeneity by profiling, and high candidate dimension by scaled Bessel evaluation. In the thin-shell regime of high dimension, amplitude shifts angular location, so profiling serves the third condition as well.

For distance calibration, the closed-form Gride KL admits the generic-order ratio, which trades a
clean-data margin for improved accuracy at every evaluated level
$\eta\geq0.1$.

For angular calibration, the local-interior and thin-shell regimes distinguish the
dimension-bearing concentration from a mean direction affected by geometry
and amplitude. Under amplitude heterogeneity ($\sigma_S=0.25$), the angular-location penalty
drives MiND--Full to $22.80\pm1.76$ against a generating dimension of $70$.
Profiling removes that penalty and returns the estimate to $66.67\pm1.81$;
dividing out the known simulated amplitudes recovers the amplitude-free
sample and its Full calibration. Full and
Profiled therefore have complementary uses: retain angular location when
references match the finite-sample geometry, and align it when location
mismatch obscures informative concentration. Making these
choices explicit lets the distance statistic and angular discrepancy be assessed separately within one combined estimate.

The image and CNN studies, reaching candidates to $400$, turn this structure into a tool
for learned representations. On CIFAR-10 and ImageNet, the Full--Profiled
separation accompanies an angular location below the reference range, and
amplitude-reducing normalization moves the location toward that range and
narrows the separation. Across four CNNs, the paired Full--Profiled comparison
locates the layers most sensitive to angular-location calibration: the largest
median separation occurs at the Profiled peak, where Full estimates are
$100$--$214$ dimensions lower.

The component structure also suggests extensions through dependence-aware weights, adaptive neighbor orders, and amplitude-aware angular references. More generally, other statistics can enter the framework once a dimension-indexed reference law and discrepancy are available. By exposing the statistical
choices within geometric calibration, the framework provides a basis for
such extensions and for adapting estimates to practical data.

\section*{CRediT authorship contribution statement}
\textbf{Chih-Hsuan Huang}: Conceptualization, Methodology, Software,
Investigation, Writing -- original draft.
\textbf{Chih-Wei Chen}: Supervision, Funding acquisition, Writing -- review \& editing.
\textbf{Szu-Chi Chung}: Conceptualization, Supervision, Funding acquisition,
Writing -- review \& editing.

\section*{Funding}
This work was supported by the National Science and Technology Council (NSTC), Taiwan, under grants NSTC 114-2118-M-110-002-MY3 and NSTC 113-2115-M-110-009-MY2. The funder had no role in the study design, data collection and analysis, interpretation of the results, preparation of the manuscript, or the decision to submit the article for publication.

\section*{Declaration of competing interest}
The authors declare that they have no competing financial interests or personal relationships that could have influenced the work reported in this paper.

\section*{Declaration of generative AI and AI-assisted technologies\\
in the manuscript preparation process}
During the preparation of this work, the authors used Claude (Anthropic) and
OpenAI Codex for language editing, for consistency and mathematical checking
of the text, and for manuscript and code organization.
After using these tools, the authors reviewed and edited the content as needed and take full responsibility
for the content of the published article.

\section*{Data and code availability}
The code supporting this study will be made publicly available on Zenodo upon acceptance of the manuscript. The datasets used in the experiments are publicly available. Synthetic benchmark datasets were generated using \texttt{scikit-dimension}~\cite{bac2021scikit}; MNIST~\cite{lecun1998gradient} and CIFAR-10~\cite{krizhevsky2009learning} were accessed through \texttt{torchvision}; and the ImageNet data~\cite{russakovsky2015imagenet} were obtained from the single-object subsets provided by~\cite{ansuini2019intrinsic}. Access instructions and processed results will be included in the Zenodo archive. Raw images are not redistributed.

\setlength{\bibsep}{2pt}
\bibliographystyle{elsarticle-num}
\bibliography{refs}

\end{document}

\maketitle

\paragraph{Notation and conventions}
We follow the notation and method definitions in the Main paper. In particular,
$d$, $D$, and $m$ denote generating, ambient, and candidate dimension;
$m_{\max}$ is the candidate bound; and $k_1<k_2$ are Gride neighbor orders.
We write $Y_d\xrightarrow{p}c$ as $d\to\infty$ when
$\Pr(|Y_d-c|>\varepsilon)\to0$ for every $\varepsilon>0$, and
$h(x)=O\{g(x)\}$ when $|h(x)|\leq C|g(x)|$ near the stated limit. The symbol
$\mathbf I_d$ denotes the $d\times d$ identity matrix. The abbreviation i.i.d.
means independent and identically distributed. The material follows the
dependencies of the Main argument: Section S1 supplies its derivations,
Section S2 connects the estimator to computation and numerical stability,
Section S3 specifies the experimental designs, and Section S4 gives the
extended evidence and component-level interpretations.

\section{Theory and derivations}
\label{app:theory}

\subsection{Gride Kullback--Leibler divergence}
\label{app:gride-kl}

Calibrating a generic-order distance statistic requires its divergence from
each candidate reference law. For neighbor orders $k_1<k_2$, define the ratio
$\mu=r_{i,k_2}/r_{i,k_1}\in(1,\infty)$. Under the local Poisson model, its
dimension-$d$ density is~\cite[Thm.~2.3, Eq.~(10)]{denti2022generalized}
\begin{equation}
 f(\mu;d,k_1,k_2)
 =\frac{d(\mu^d-1)^{k_2-k_1-1}}
 {\mu^{d(k_2-1)+1}B(k_2-k_1,k_1)},\qquad \mu>1,
 \label{eq:supp-gride-density}
\end{equation}
where
$B(\alpha,\beta)=\Gamma(\alpha)\Gamma(\beta)/
\Gamma(\alpha+\beta)$ is the beta function~\cite[Secs.~5.2 and 5.12]{olver2010nist}.

Let $d_1=\dhat_{\mathrm{dist}}$ be the fitted data dimension and
$d_2=\hat d_m^{\mathrm{ref}}$ the fitted dimension of the candidate-$m$
reference. The two laws compared by calibration are
\[
 f(\mu)=f(\mu;d_1,k_1,k_2),\qquad
 f_m(\mu)=f(\mu;d_2,k_1,k_2).
\]
Their KL divergence is
\[
 \KL(f\|f_m)
 =\mathbb E_f\!\left[\log\frac{f(\mu)}{f_m(\mu)}\right].
\]
Taking logarithms in Eq.~\eqref{eq:supp-gride-density} gives
\begin{align*}
 \log f(\mu)
 ={}&\log d_1+(k_2-k_1-1)\log(\mu^{d_1}-1)\\
 &-[d_1(k_2-1)+1]\log\mu-\log B(k_2-k_1,k_1),\\
 \log f_m(\mu)
 ={}&\log d_2+(k_2-k_1-1)\log(\mu^{d_2}-1)\\
 &-[d_2(k_2-1)+1]\log\mu-\log B(k_2-k_1,k_1).
\end{align*}
The common beta normalizer cancels. Subtracting and taking expectation yields
\begin{align}
 \KL(f\|f_m)
 ={}&\log\frac{d_1}{d_2}
 +(k_2-1)(d_2-d_1)\mathbb E_f[\log\mu] \notag\\
 &+(k_2-k_1-1)\left\{
 \mathbb E_f[\log(\mu^{d_1}-1)]
 -\mathbb E_f[\log(\mu^{d_2}-1)]\right\}.
 \label{eq:supp-decomp}
\end{align}

Equation~\eqref{eq:supp-decomp} separates the required divergence into three
expectations. The following lemma evaluates two beta-prime log moments in
closed form, leaving one reference-side expectation, evaluated below in closed form or by one-dimensional quadrature.

\begin{lemma}[Beta-prime log moments]
\label{lem:betaprime}
If $\alpha,\beta>0$ and $z\sim\beta'(\alpha,\beta)$ with density
$z^{\alpha-1}(1+z)^{-\alpha-\beta}/B(\alpha,\beta)$, then
\[
\mathbb E[\log(1+z)]=\Psi(\alpha+\beta)-\Psi(\beta),\qquad
\mathbb E[\log z]=\Psi(\alpha)-\Psi(\beta),
\]
where $\Gamma$ is the gamma function and
\[
 B(\alpha,\beta)
 =\frac{\Gamma(\alpha)\Gamma(\beta)}{\Gamma(\alpha+\beta)},
 \qquad
 \Psi(x)=\frac{d}{dx}\log\Gamma(x)
\]
define the beta and digamma functions~\cite[Secs.~5.2 and 5.12]{olver2010nist}.
\end{lemma}

\begin{proof}
The beta-prime normalizer is
\[
 B(\alpha,\beta)=\int_0^\infty
 z^{\alpha-1}(1+z)^{-\alpha-\beta}\,dz.
\]
Near zero, the kernel multiplied by the logarithmic factors is of order
$z^{\alpha-1}|\log z|$; near infinity it is of order
$z^{-\beta-1}\log z$. Both are integrable for $\alpha,\beta>0$. The same bounds
hold uniformly on a small closed neighborhood of $(\alpha,\beta)$, so the
derivative may pass under the integral sign. Writing $\partial_\beta$ for the
derivative with respect to $\beta$ while $\alpha$ is held fixed,
differentiation and division by $B(\alpha,\beta)$ give
\[
 \partial_\beta\log B(\alpha,\beta)
 =-\mathbb E[\log(1+z)]
 =\Psi(\beta)-\Psi(\alpha+\beta).
\]
Differentiating while holding $\beta$ fixed similarly gives
\[
 \partial_\alpha\log B(\alpha,\beta)
 =\mathbb E[\log z-\log(1+z)]
 =\Psi(\alpha)-\Psi(\alpha+\beta).
\]
The first identity yields the stated expectation of $\log(1+z)$; substituting
it into the second yields the expectation of $\log z$.
\end{proof}

To evaluate the three expectations in Eq.~\eqref{eq:supp-decomp}, transform
$z=\mu^{d_1}-1$. The inverse and Jacobian are
\[
 \mu=(1+z)^{1/d_1},\qquad
 \frac{d\mu}{dz}=\frac{1}{d_1}(1+z)^{1/d_1-1}.
\]
The change-of-variables formula gives the transformed density explicitly:
\[
\begin{split}
 p_Z(z)
 &=f\bigl((1+z)^{1/d_1};d_1,k_1,k_2\bigr)
   \left|\frac{d\mu}{dz}\right|\\
 &=\frac{z^{k_2-k_1-1}}
 {B(k_2-k_1,k_1)}
 (1+z)^{-\{d_1(k_2-1)+1\}/d_1+1/d_1-1}\\
 &=\frac{z^{k_2-k_1-1}(1+z)^{-k_2}}
 {B(k_2-k_1,k_1)},\qquad z>0.
\end{split}
\]
Thus $z\sim\beta'(k_2-k_1,k_1)$, as also noted in the Supplementary Material of Denti et al.~\cite{denti2022generalized}, and Lemma~\ref{lem:betaprime} gives
\begin{equation}
 \mathbb E_f[\log\mu]
  =\frac{\Psi(k_2)-\Psi(k_1)}{d_1},
 \label{eq:supp-e1}
\end{equation}
and
\begin{equation}
 \mathbb E_f[
    \log(\mu^{d_1}-1)]
  =\Psi(k_2-k_1)-\Psi(k_1).
 \label{eq:supp-e2}
\end{equation}
The reference-side term uses the same substitution. Let $\gamma_m=d_2/d_1>0$ denote the ratio of the fitted reference dimension to the fitted data dimension, so that $d_2=\gamma_m d_1$; the two fitted distance laws coincide exactly when $\gamma_m=1$. Since $\mu=(1+z)^{1/d_1}$,
\[
\begin{split}
 \log(\mu^{d_2}-1)
 &=\log\!\left\{\left((1+z)^{1/d_1}\right)^{\gamma_m d_1}-1\right\}\\
 &=\log\{(1+z)^{\gamma_m}-1\}.
\end{split}
\]
For $\gamma_m\neq1$ this is neither of the two log moments of Lemma~\ref{lem:betaprime}, so its expectation under $z\sim\beta'(k_2-k_1,k_1)$, denoted $\mathcal I(\gamma_m)$, has to be evaluated separately. By definition,
\begin{equation}
  \mathcal I(\gamma_m)
  =\int_0^\infty
  \frac{z^{k_2-k_1-1}(1+z)^{-k_2}}{B(k_2-k_1,k_1)}
  \log((1+z)^{\gamma_m}-1)\,dz.
  \label{eq:supp-I}
\end{equation}
For the integer neighbor orders used throughout, $\mathcal I(\gamma_m)$ also has a closed form. The derivation moves the integral to the unit interval, splits the logarithm into a known moment and a finite binomial sum, and evaluates each summand with a second digamma identity. Transform $T=(1+z)^{-1}=\mu^{-d_1}\in(0,1)$. The inverse and Jacobian are
\[
 z=T^{-1}-1,\qquad
 \left|\frac{dz}{dT}\right|=T^{-2},
\]
so the change-of-variables formula applied to $p_Z$ gives
\[
\begin{split}
 p_T(t)
 &=p_Z(t^{-1}-1)\,t^{-2}
 =\frac{(t^{-1}-1)^{k_2-k_1-1}\,t^{k_2}\,t^{-2}}{B(k_2-k_1,k_1)}\\
 &=\frac{t^{k_1-1}(1-t)^{k_2-k_1-1}}{B(k_2-k_1,k_1)},
 \qquad 0<t<1,
\end{split}
\]
because $(t^{-1}-1)^{k_2-k_1-1}=t^{-(k_2-k_1-1)}(1-t)^{k_2-k_1-1}$ and the exponents of $t$ add to $k_1-1$. Thus $T\sim\mathrm{Beta}(k_1,k_2-k_1)$, with the same normalizer since $B(\alpha,\beta)=B(\beta,\alpha)$. In the new variable, $(1+z)^{\gamma_m}-1=T^{-\gamma_m}(1-T^{\gamma_m})$, so the logarithm splits into two terms, each absolutely integrable under $p_T$: near $t=0$ the density-weighted magnitudes are of order $t^{k_1-1}|\log t|$ and $t^{k_1+\gamma_m-1}$, and near $t=1$ of order $(1-t)^{k_2-k_1}$ and $(1-t)^{k_2-k_1-1}|\log(1-t)|$. Hence
\begin{equation}
 \mathcal I(\gamma_m)
 =\gamma_m\,\mathbb E[-\log T]+\mathbb E[\log(1-T^{\gamma_m})].
 \label{eq:supp-I-split}
\end{equation}
The first expectation is already known: $-\log T=\log(1+z)$, so Lemma~\ref{lem:betaprime} with $(\alpha,\beta)=(k_2-k_1,k_1)$ gives $\mathbb E[-\log T]=\Psi(k_2)-\Psi(k_1)$, the moment behind Eq.~\eqref{eq:supp-e1}. For the second, write $n=k_2-k_1-1$. Because $n$ is a nonnegative integer, the factor $(1-t)^{n}$ of the density is a finite binomial sum, $(1-t)^{n}=\sum_{j=0}^{n}(-1)^j\binom{n}{j}t^{j}$, and each term multiplies $t^{k_1-1}$ to a pure power. Hence
\begin{equation}
 \mathbb E[\log(1-T^{\gamma_m})]
 =\frac{1}{B(k_2-k_1,k_1)}\sum_{j=0}^{n}(-1)^j\binom{n}{j}\,\mathcal J(k_1+j,\gamma_m),
 \qquad
 \mathcal J(s,\gamma)=\int_0^1 t^{s-1}\log(1-t^{\gamma})\,dt.
 \label{eq:supp-J-def}
\end{equation}
To evaluate $\mathcal J$, substitute $u=t^{\gamma}$, so that $t=u^{1/\gamma}$ and $dt=\gamma^{-1}u^{1/\gamma-1}\,du$:
\[
 \mathcal J(s,\gamma)=\frac{1}{\gamma}\int_0^1 u^{s/\gamma-1}\log(1-u)\,du.
\]
The remaining integral is a beta log moment of the type treated in Lemma~\ref{lem:betaprime}. Differentiating $B(a,b)=\int_0^1u^{a-1}(1-u)^{b-1}\,du$ with respect to $b$, under the integral sign by the same domination argument as in the lemma, gives $\int_0^1u^{a-1}(1-u)^{b-1}\log(1-u)\,du=B(a,b)\{\Psi(b)-\Psi(a+b)\}$, and setting $b=1$, where $B(a,1)=1/a$, yields
\[
 \int_0^1 u^{a-1}\log(1-u)\,du=\frac{\Psi(1)-\Psi(1+a)}{a},\qquad a>0.
\]
With $a=s/\gamma$ the prefactor $1/\gamma$ cancels against $\gamma/s$, so
\begin{equation}
 \mathcal J(s,\gamma)=\frac{\Psi(1)-\Psi(1+s/\gamma)}{s}
 =\frac{\Psi(1)-\Psi(s/\gamma)}{s}-\frac{\gamma}{s^{2}},
 \label{eq:supp-J}
\end{equation}
where the second form uses the digamma recurrence $\Psi(1+x)=\Psi(x)+1/x$ at $x=s/\gamma$. Inserting Eq.~\eqref{eq:supp-J} with $s=k_1+j$ into Eq.~\eqref{eq:supp-J-def} produces two sums. The second one is the same binomial expansion applied to the first expectation in Eq.~\eqref{eq:supp-I-split}: since $\int_0^1t^{s-1}(-\log t)\,dt=1/s^{2}$,
\[
 -\frac{\gamma_m}{B(k_2-k_1,k_1)}\sum_{j=0}^{n}\frac{(-1)^j\binom{n}{j}}{(k_1+j)^{2}}
 =-\frac{\gamma_m}{B(k_2-k_1,k_1)}\int_0^1 t^{k_1-1}(1-t)^{n}(-\log t)\,dt
 =-\gamma_m\,\mathbb E[-\log T].
\]
This second sum cancels the first term of Eq.~\eqref{eq:supp-I-split} exactly, and only the first sum survives:
\begin{equation}
 \mathcal I(\gamma_m)
 =\frac{1}{B(k_2-k_1,k_1)}\sum_{j=0}^{k_2-k_1-1}
 \frac{(-1)^j\binom{k_2-k_1-1}{j}}{k_1+j}
 \left\{\Psi(1)-\Psi\!\left(\frac{k_1+j}{\gamma_m}\right)\right\}.
 \label{eq:supp-I-closed}
\end{equation}
The released implementation evaluates Eq.~\eqref{eq:supp-I} by adaptive one-dimensional quadrature; over the candidate ranges used in this study ($m\le100$ for $k\in\{5,15,20\}$ and $m\le400$ for $k=10$) the two evaluations agree to within $10^{-6}$. Substituting Eqs.~\eqref{eq:supp-e1}--\eqref{eq:supp-I} into
Eq.~\eqref{eq:supp-decomp} gives the complete calibrated divergence:
\begin{equation}
\begin{split}
 \KL(f\|f_m)
 ={}&\log\frac{\dhat_{\mathrm{dist}}}{\hat d_m^{\mathrm{ref}}}
 +(k_2-1)\left(\frac{\hat d_m^{\mathrm{ref}}}
 {\dhat_{\mathrm{dist}}}-1\right)
 \bigl\{\Psi(k_2)-\Psi(k_1)\bigr\}\\
 &+(k_2-k_1-1)\left\{\Psi(k_2-k_1)-\Psi(k_1)
 -\mathcal I(\gamma_m)\right\}.
\end{split}
\label{eq:supp-klgride}
\end{equation}
Equation~\eqref{eq:supp-klgride} gives Main Eq.~(7), with $\mathcal I(\gamma_m)$ given by Eq.~\eqref{eq:supp-I-closed} or by the quadrature of Eq.~\eqref{eq:supp-I}.
For the matching-law consistency check,
$\gamma_m=1$ gives
$\mathcal I(1)=\mathbb E[\log z]=\Psi(k_2-k_1)-\Psi(k_1)$, so every term
vanishes and $\KL(f\|f)=0$.

\subsection{Angular limits and the amplitude mechanism}
\label{app:nu-proofs}

The angular analysis separates local symmetry from fixed-sample
high-dimensional geometry. Across these two regimes, the mean direction of
neighbor angles has three limiting descriptions: the dimension-free $\pi/2$
of the local-interior regime
(Proposition~\ref{prop:supp-nu-symmetric}), the value $\pi/3$ reached at fixed
$N$ as $d\to\infty$ under a common amplitude
(Lemma~\ref{lem:supp-thin-shell}, Proposition~\ref{prop:supp-shell-limit}, and
Corollaries~\ref{cor:supp-homogeneous-shell} and~\ref{cor:supp-uniform-ball}),
and an amplitude-dependent limit below $\pi/3$ when both neighbor amplitudes
are smaller than the center amplitude
(Corollary~\ref{cor:supp-gsm-shell}).

Proposition~\ref{prop:supp-nu-symmetric} uses the following local-symmetry
condition. At an interior point of a smooth
manifold~\cite[Secs.~2.1 and 3.1]{meila2024manifold},
we assume that the sampling density is continuous and
positive, that the kNN radius shrinks, and that the normalized neighbor
displacements converge to independent uniform directions on the unit sphere of
the tangent space. Because $\theta$, $\sin\theta$, and $\cos\theta$ are bounded and continuous, this convergence carries the arithmetic and circular means over to the limit law. This is the manifold condition needed to apply the
statement below, which is Main Proposition 1.

\begin{proposition}[Dimension-free mean direction under local symmetry]
\label{prop:supp-nu-symmetric}
Let $u,v$ be independent uniform directions in $\mathbb R^d$, and let
$\theta$ be their angle. Its probability density is
\begin{equation}
  h_d(\theta)
  =\frac{\Gamma(d/2)}
  {\sqrt{\pi}\Gamma((d-1)/2)}
  \sin^{d-2}\theta,\qquad 0\leq\theta\leq\pi.
  \label{eq:supp-angle-density}
\end{equation}
Here $\Gamma$ is the gamma function. The density is symmetric about $\pi/2$,
so the arithmetic mean and circular mean direction both equal $\pi/2$ for
every $d\geq2$.
\end{proposition}

\begin{proof}
The angle density follows from rotational invariance and the standard law of
the inner product of two random directions~\cite[Lemmas~11--12, Eqs.~(12)--(13)]{cai2013distributions}. Because
$h_d(\pi-\theta)=h_d(\theta)$,
\[
 \mathbb E(\theta)-\frac{\pi}{2}
 =\int_0^\pi\left(\theta-\frac{\pi}{2}\right)h_d(\theta)\,d\theta=0.
\]
The cosine is antisymmetric about $\pi/2$, whereas the sine is positive on
$(0,\pi)$, so
\[
 \mathbb E(\cos\theta)=0,
 \qquad
 \mathbb E(\sin\theta)=\int_0^\pi\sin\theta\,h_d(\theta)\,d\theta>0.
\]
The circular mean of Main Section 3.1 therefore gives
\[
 \nu=\atantwo\{\mathbb E(\sin\theta),\mathbb E(\cos\theta)\}
 =\atantwo(>0,0)=\frac{\pi}{2}.
\]
\end{proof}

Cai et al. further show, in their growing-sample, growing-dimension regime, that
the empirical distribution of $\sqrt{d-2}(\pi/2-\theta)$ converges to the
standard normal law, so the unscaled angle deviations have a $d^{-1/2}$ scale
around $\pi/2$~\cite[Thm.~4]{cai2013distributions}. Original
DANCo combines this limit with the large-concentration von Mises approximation:
under the ideal symmetric model, $\nu=\pi/2$ while $\tau/d\to1$~\cite[Sec.~3.2, Prop.~1]{ceruti2014danco}. ABID expresses the same
dimension-dependent spread through
$\mathbb E(\cos^2\theta)=1/d$~\cite[Cor.~1]{thordsen2022abid}.

The fixed-sample, increasing-dimension regular-simplex geometry is classical~\cite[Sec.~3.1]{hall2005geometric}. The following i.i.d.-coordinate lemma is a
direct specialization to the norm and cross-inner-product limits used here.

\begin{lemma}[Thin-shell concentration and near-orthogonality]
\label{lem:supp-thin-shell}
For each $s$ in a fixed finite index set, let
$Z_{s,1},Z_{s,2},\ldots$ be an i.i.d. sequence with mean zero, variance one,
and finite fourth moment. Assume the sequences are mutually independent and let $Z_s^{(d)}=(Z_{s,1},\ldots,Z_{s,d})\in\mathbb R^{d}$ be the vector of its first $d$ coordinates. Then
\[
 \frac{\|Z_s^{(d)}\|^2}{d}\xrightarrow{p}1,
 \qquad
 \frac{\langle Z_s^{(d)},Z_t^{(d)}\rangle}{d}\xrightarrow{p}0
 \quad(s\ne t)
\]
simultaneously over the fixed collection.
\end{lemma}

\begin{proof}
Fix $s$ and put $Y_r=Z_{s,r}^2$. Then
$\mathbb E(Y_r)=\mathrm{Var}(Z_{s,r})+\{\mathbb E(Z_{s,r})\}^2=1+0=1$ and
$\mathrm{Var}(Y_r)=\mathbb E(Z_{s,r}^4)-1$, a finite constant; this is the only
place where the finite fourth moment is used. The $Y_r$ are therefore i.i.d.
with finite variance, and the weak law of large numbers~\cite[Thm.~2.4]{blum2020foundations} gives
$d^{-1}\sum_{r=1}^d Y_r\xrightarrow{p}1$, which is the first limit.

For $s\ne t$ put $W_r=Z_{s,r}Z_{t,r}$. The two sequences are mutually
independent, so the $W_r$ are i.i.d. with
$\mathbb E(W_r)=\mathbb E(Z_{s,r})\mathbb E(Z_{t,r})=0\cdot0=0$ and
$\mathrm{Var}(W_r)=\mathbb E(Z_{s,r}^2)\mathbb E(Z_{t,r}^2)=1$. The same
theorem gives $d^{-1}\sum_{r=1}^d W_r\xrightarrow{p}0$, which is the second
limit.

Both limits hold simultaneously over the fixed collection. That collection
indexes finitely many events, one per norm and one per pair. Fix
$\varepsilon>0$ and let $E_d$ be the event that at least one of them
deviates from its limit by more than $\varepsilon$; then $\Pr(E_d)$
is at most the sum of the finitely many individual probabilities, each of which
tends to zero, so $\Pr(E_d)\to0$.
\end{proof}

To translate these limits into displacement angles, the next proof uses an
elementary consequence of convergence in probability. If finitely many
quantities converge jointly to constants, then
their sums, products, and any continuous function of them also converge. A
ratio may be included when its limiting denominator is nonzero; this is the
continuous-mapping theorem and the special case of Slutsky's theorem needed
below. The statement that follows is Main Proposition 2.

\begin{proposition}[Displacement-angle limit under norm and inner-product concentration]
\label{prop:supp-shell-limit}
Fix $N$ and $k$ while $d\to\infty$. For each $d$, let
$X_s^{(d)}\in\mathbb R^d$, $s\in\{i,j,\ell\}$, and suppose that there are
nonnegative constants $R_s$ such that, simultaneously,
\[
  \|X_s^{(d)}\|^2/d\xrightarrow{p}R_s^2,\qquad
  \langle X_s^{(d)},X_t^{(d)}\rangle/d\xrightarrow{p}0\quad(s\ne t).
\]
Assume additionally that
$(R_i^2+R_j^2)(R_i^2+R_\ell^2)>0$.
Then
\begin{equation}
  \cos\theta_{j\ell}^{(i)}
  \xrightarrow{p}
  \frac{R_i^2}
  {\sqrt{(R_i^2+R_j^2)(R_i^2+R_\ell^2)}}.
  \label{eq:supp-shell-angle-limit}
\end{equation}
If the assumptions hold simultaneously for all observations, the convergence
is simultaneous over all selected triples.
\end{proposition}

\begin{proof}
Write the cosine as a normalized numerator divided by two normalized
displacement lengths:
\[
 \cos\theta_{j\ell}^{(i)}
 =\frac{A_d}{(B_{j,d}B_{\ell,d})^{1/2}},
 \quad
 A_d=\frac{\langle X_j^{(d)}-X_i^{(d)},
 X_\ell^{(d)}-X_i^{(d)}\rangle}{d},
\]
where
\[
 B_{j,d}=\frac{\|X_j^{(d)}-X_i^{(d)}\|^2}{d},\qquad
 B_{\ell,d}=\frac{\|X_\ell^{(d)}-X_i^{(d)}\|^2}{d}.
\]
Expanding all four inner-product terms in $A_d$ gives
\[
 A_d=\frac{\langle X_j^{(d)},X_\ell^{(d)}\rangle
 -\langle X_j^{(d)},X_i^{(d)}\rangle
 -\langle X_\ell^{(d)},X_i^{(d)}\rangle
 +\|X_i^{(d)}\|^2}{d}.
\]
The three cross terms vanish in the limit by the inner-product assumption and
the last term converges to $R_i^2$ by the norm assumption, so
$A_d\xrightarrow{p}R_i^2$. The same expansion for the squared lengths leaves
two norm terms and one cross term in each, whence
$B_{j,d}\xrightarrow{p}R_j^2+R_i^2$ and
$B_{\ell,d}\xrightarrow{p}R_\ell^2+R_i^2$. The positivity assumption
$(R_i^2+R_j^2)(R_i^2+R_\ell^2)>0$ makes the limiting denominator strictly
positive, so the square root and the division are continuous at the limit
point; the continuous-mapping theorem and Slutsky's theorem then give
Eq.~\eqref{eq:supp-shell-angle-limit}. Simultaneity over the finitely many
possible triples, and hence over any selected subset, follows from the same
finite union bound as in Lemma~\ref{lem:supp-thin-shell}.
\end{proof}

\begin{corollary}[Homogeneous thin shell]
\label{cor:supp-homogeneous-shell}
Suppose each observation consists of the first $d$ coordinates of an infinite
i.i.d. sequence with mean zero, variance $\sigma^2>0$, and finite fourth
moment, and the sequences are mutually independent. Then
\[
 \frac{\|X_s^{(d)}\|^2}{d}\xrightarrow{p}\sigma^2,\qquad
 \frac{\langle X_s^{(d)},X_t^{(d)}\rangle}{d}\xrightarrow{p}0,
\]
and every selected triple satisfies
\[
 \cos\theta_{j\ell}^{(i)}\xrightarrow{p}\frac12,\qquad
 \theta_{j\ell}^{(i)}\xrightarrow{p}\frac{\pi}{3}.
\]
\end{corollary}

\begin{proof}
The coordinates of $X_s^{(d)}/\sigma$ are i.i.d. with mean zero, variance one,
and finite fourth moment, so Lemma~\ref{lem:supp-thin-shell} applies to them;
multiplying back by $\sigma$ gives the stated norm and inner-product limits.
Substituting $R_i=R_j=R_\ell=\sigma$ in
Proposition~\ref{prop:supp-shell-limit} gives
$\cos\theta_{j\ell}^{(i)}\xrightarrow{p}1/2$, and continuity of $\arccos$ at
$1/2$ gives the angle limit.
\end{proof}

\begin{corollary}[Uniform hyperball]
\label{cor:supp-uniform-ball}
Let $X_1^{(d)},\ldots,X_N^{(d)}$ be independent and uniformly distributed in
the $d$-dimensional unit ball. For fixed $N$, all selected-neighbor angles
converge in probability to $\pi/3$ as $d\to\infty$.
\end{corollary}

\begin{proof}
Write $X_s^{(d)}=\rho_s^{(d)}U_s^{(d)}$, where $\rho_s^{(d)}$ and
$U_s^{(d)}$ are its radius and direction. The ball of radius
$1-\varepsilon$ occupies the fraction $(1-\varepsilon)^d$ of the volume of the
unit ball, so uniform-ball sampling
gives $\Pr\{\rho_s^{(d)}\leq1-\varepsilon\}=(1-\varepsilon)^d\to0$ for every
$\varepsilon\in(0,1)$~\cite[Sec.~2.3]{blum2020foundations}; since
$\rho_s^{(d)}\leq1$, this is exactly $\rho_s^{(d)}\xrightarrow{p}1$ and the
mass collects on the boundary shell. Independent uniform directions satisfy
$\langle U_s^{(d)},U_t^{(d)}\rangle\xrightarrow{p}0$~\cite[Lemmas~11--12]{cai2013distributions}. Hence
$\|X_s^{(d)}\|^2\xrightarrow{p}1$ and
$\langle X_s^{(d)},X_t^{(d)}\rangle\xrightarrow{p}0$. Rescaling all points by
the common factor $\sqrt d$ leaves every displacement angle unchanged, so
Proposition~\ref{prop:supp-shell-limit} may be applied to
$\widetilde X_s^{(d)}=\sqrt d\,X_s^{(d)}$, which satisfies its hypotheses with
$R_i=R_j=R_\ell=1$ and gives
\[
 \cos\theta_{j\ell}^{(i)}\xrightarrow{p}\frac12,\qquad
 \theta_{j\ell}^{(i)}\xrightarrow{p}\frac{\pi}{3}.
\]
\end{proof}

\begin{corollary}[Gaussian scale mixture]
\label{cor:supp-gsm-shell}
Let $X_s^{(d)}=S_sZ_s^{(d)}$, where the
$Z_s^{(d)}\sim\mathcal N(0,\mathbf I_d)$ are independent and
the scale vector $(S_1,\ldots,S_N)$ is independent of all Gaussian vectors.
Conditional on the positive amplitudes,
\[
  \|X_s^{(d)}\|^2/d\xrightarrow{p}S_s^2,\qquad
  \langle X_s^{(d)},X_t^{(d)}\rangle/d\xrightarrow{p}0.
\]
For selected neighbors with amplitudes $S_j$ and $S_\ell$,
\[
 \frac{\|X_j^{(d)}-X_i^{(d)}\|^2}{d}
 \xrightarrow{p}S_i^2+S_j^2,
\]
and
\[
 \cos\theta_{j\ell}^{(i)}\xrightarrow{p}
 \frac{S_i^2}{\sqrt{(S_i^2+S_j^2)(S_i^2+S_\ell^2)}}.
\]
If $S_j<S_i$ and $S_\ell<S_i$, the limiting angle is below $\pi/3$, including
the case $S_j\ne S_\ell$.
\end{corollary}

\begin{proof}
Condition on the positive amplitudes. The Gaussian vectors remain independent
with i.i.d. standard normal coordinates, whose fourth moment is finite, so
Lemma~\ref{lem:supp-thin-shell} applies to them; multiplying by the
conditionally constant $S_s$ scales the norm limit to $S_s^2$ and leaves the
inner-product limit at zero. Expanding the displacement length gives the
distance limit, and Proposition~\ref{prop:supp-shell-limit} with $R_s=S_s$
gives the cosine; the positive amplitudes make
$(S_i^2+S_j^2)(S_i^2+S_\ell^2)>0$, so its positivity assumption holds. If
both selected
amplitudes are below $S_i$, each denominator factor is below $2S_i^2$; the
limiting cosine exceeds $1/2$, and the decreasing function $\arccos$ gives an
angle below $\pi/3$.
\end{proof}

For the lognormal GSM used in the finite-sample experiments,
$S_s>0$ are independent with
$\log S_s\sim\mathcal N(0,\sigma_S^2)$, independently of the Gaussian vectors.
Thus $\sigma_S$ is the standard deviation of log amplitude: $\sigma_S=0$
gives a common amplitude, while larger values increase sample-amplitude
heterogeneity without changing the generating dimension.

\begin{figure}[ht]
\centering
\includegraphics[width=0.62\linewidth]{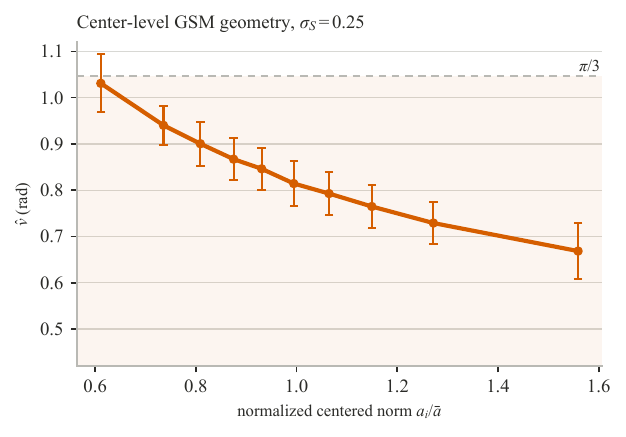}
\caption{Normalized centered norm $a_i/\bar a$ and per-center mean direction within a $d=70$, $\sigma_S=0.25$ GSM sample, with $\bar a$ the mean centered norm. Deciles contain equal numbers of centers;
error bars show within-decile standard deviations, and the dashed horizontal
line marks $\pi/3$.}
\label{fig:nu-mechanism-supp}
\end{figure}

\paragraph{Purpose and construction of the finite-sample diagnostics}
The finite-sample diagnostic uses centered norm $a_i=\|x_i-\bar x\|_2$, with $\bar x=N^{-1}\sum_i x_i$, as the observable measure of an observation's amplitude, and reports it relative to its sample mean $\bar a=N^{-1}\sum_i a_i$. By Corollary~\ref{cor:supp-gsm-shell}, the limiting cosine increases with the center amplitude relative to its selected neighbors, so a larger centered norm, the observable proxy for that amplitude, is expected to go with a smaller per-center mean direction. Figure~\ref{fig:nu-mechanism-supp} bins centered norm into deciles in a separate GSM sample at the focal level $\sigma_S=0.25$ of Section~\ref{app:design-gsm} and reports the mean and within-decile standard deviation of $\hat\nu_i$; the decile means decrease monotonically from $1.031$ to $0.668$ radians. Main Figure~2 examines the finite-$N$ transition between angular regimes using the design in Section~\ref{app:design-gsm}.

\subsection{Profiled circular von Mises divergence}
\label{app:profiled-divergence}

The circular von Mises KL divergence in Main Eq.~(4), taken from Vo et al.~\cite{vo2008statistical}, is
\[
 \KL\{q(\cdot;\nu_1,\tau_1)\,\|\,q(\cdot;\nu_2,\tau_2)\}
 =\log\frac{I_0(\tau_2)}{I_0(\tau_1)}
 +A(\tau_1)\{\tau_1-\tau_2\cos(\nu_2-\nu_1)\},
 \qquad A(\tau)=\frac{I_1(\tau)}{I_0(\tau)}.
\]
Adding and subtracting $A(\tau_1)\tau_2$ gives the decomposition stated after Main Eq.~(4): a concentration term plus an angular-location penalty,
\[
 \KL\{q_1\|q_2\}
 =\underbrace{\log\frac{I_0(\tau_2)}{I_0(\tau_1)}+A(\tau_1)(\tau_1-\tau_2)}_{\text{concentration}}
 +\underbrace{A(\tau_1)\tau_2\{1-\cos(\nu_2-\nu_1)\}}_{\text{location penalty}}.
\]
The previous subsection shows that sample-amplitude heterogeneity can move the mean direction without changing the generating dimension, so the relative mean direction is treated as a nuisance parameter. Following the general rule of Main Eq.~(6), let $\delta$ rotate the reference mean direction; the profiled angular discrepancy of Main Eq.~(9) substitutes $(\nu_1,\tau_1)=(\hat\nu,\hat\tau)$ and $(\nu_2,\tau_2)=(\nu_m^{\mathrm{ref}}+\delta,\tau_m^{\mathrm{ref}})$ and takes the infimum over $\delta$:
\[
 \Delta_{\mathrm{ang}}^{\mathrm{prof}}(m)
 =\inf_{\delta}\Bigl[
 \log\frac{I_0(\tau_m^{\mathrm{ref}})}{I_0(\hat\tau)}
 +A(\hat\tau)(\hat\tau-\tau_m^{\mathrm{ref}})
 +A(\hat\tau)\tau_m^{\mathrm{ref}}\{1-\cos(\nu_m^{\mathrm{ref}}+\delta-\hat\nu)\}
 \Bigr].
\]
Only the last term depends on $\delta$. For $\tau\geq0$ the integral representation gives $I_0(\tau)>0$ and $I_1(\tau)\geq0$, so $A(\hat\tau)\tau_m^{\mathrm{ref}}\geq0$, and $1-\cos x\geq0$ makes the term nonnegative. It attains its minimum value zero when the mean directions are aligned, $\nu_m^{\mathrm{ref}}+\delta=\hat\nu$ modulo $2\pi$; if $A(\hat\tau)\tau_m^{\mathrm{ref}}=0$, the bracket does not depend on $\delta$ and the same value results. Therefore
\[
 \Delta_{\mathrm{ang}}^{\mathrm{prof}}(m)
 =\log\frac{I_0(\tau_m^{\mathrm{ref}})}{I_0(\hat\tau)}
 +A(\hat\tau)(\hat\tau-\tau_m^{\mathrm{ref}}),
\]
which is Main Eq.~(10): the infimum is attained by aligning the mean directions, and only the concentration term remains.

Full matching keeps the original reference mean direction, so its angular discrepancy $\Delta_{\mathrm{ang}}(m)$ in Main Eq.~(5) is Main Eq.~(4) at $\delta=0$, and its excess over the profiled discrepancy is
\[
 \Delta_{\mathrm{ang}}(m)-\Delta_{\mathrm{ang}}^{\mathrm{prof}}(m)
 =A(\hat\tau)\tau_m^{\mathrm{ref}}\{1-\cos(\nu_m^{\mathrm{ref}}-\hat\nu)\}\geq0.
\]
Profiling therefore removes exactly the angular-location penalty of Main Eq.~(4) while retaining the concentration comparison. For a fixed offset that is not a multiple of $2\pi$ and $\hat\tau>0$, this penalty increases with $\tau_m^{\mathrm{ref}}$; when both the offset and the reference concentration vary across candidates, its candidate-wise behavior depends on both quantities.

\subsection{Adequacy of the circular von Mises approximation}
\label{app:model-adequacy}

Under local symmetry the angle density is exactly
$h_d(\theta)\propto\sin^{d-2}\theta$ on $[0,\pi]$
(Proposition~\ref{prop:supp-nu-symmetric},
Eq.~\eqref{eq:supp-angle-density}). The exact density is therefore the locally symmetric benchmark. Its symmetry about $\pi/2$ also fixes its mean direction
there for every $d$, so it has no parameter with which to express the
data--reference difference in angular location that the observed samples
display.

DANCo accommodates variable angular location through the von Mises family:
it represents each observed and reference mutual-neighbor angle distribution
with a fitted circular von Mises
approximation~\cite[Sec.~3.2]{ceruti2014danco}, and that family is the standard
parametric model for circular data~\cite[Sec.~3.3.6]{fisher1993statistical}.
Writing $q_{\mathrm{circ}}$ for the density $q(\cdot;\nu,\tau)$ of Main
Section 3.1, we separately fit this law by maximum likelihood to the pooled
neighbor angles and assess its descriptive fit. These parameters differ from
the aggregated per-center statistics used for calibration.

Because the physical angles occupy $[0,\pi]$ while $q_{\mathrm{circ}}$ is
circular, we first measure its outside-range mass,
\begin{equation}
 \epsilon(\nu,\tau)
 =1-\int_0^\pi q_{\mathrm{circ}}(\theta;\nu,\tau)\,d\theta
 \label{eq:supp-vm-outside}
\end{equation}
and then assess its fit on the physical range with the renormalized density
\begin{equation}
 q_{[0,\pi]}(\theta;\nu,\tau)
 =\frac{q_{\mathrm{circ}}(\theta;\nu,\tau)}
 {1-\epsilon(\nu,\tau)}\,\mathbf 1_{[0,\pi]}(\theta).
 \label{eq:supp-vm-truncated}
\end{equation}
For these fitted pooled models, Eq.~\eqref{eq:supp-vm-outside} gives
$\epsilon$ at most $5.0\times10^{-8}$ and falls
below $10^{-8}$ on every real dataset, so the circular approximation places
essentially all fitted mass in the physical range.

\begin{figure}[ht]
\centering
\includegraphics[width=0.95\linewidth]{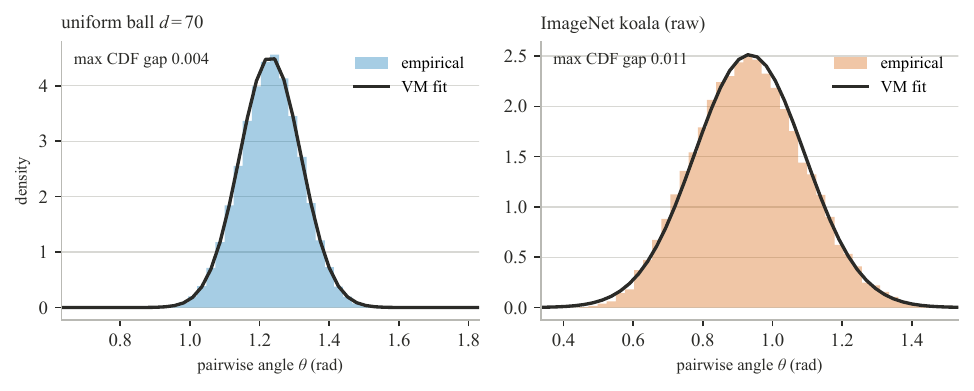}
\caption{Representative pooled neighbor-angle distributions and fitted von Mises (VM) laws for the uniform 70-ball and raw ImageNet koala. The koala fit has a
modest descriptive CDF gap; its dominant data--reference discrepancy occurs in
angular location.}
\label{fig:vm-adequacy}
\end{figure}

For the pooled $M=N\binom{k}{2}$ angles, let $Q$ denote the cumulative distribution function (CDF) of
Eq.~\eqref{eq:supp-vm-truncated} and let
$\theta_{(1)}\leq\cdots\leq\theta_{(M)}$ be the ordered observations. We report
\[
 D^+=\max_r\left\{\frac rM-Q(\theta_{(r)})\right\},\qquad
 D^-=\max_r\left\{Q(\theta_{(r)})-\frac{r-1}{M}\right\},
 \qquad G=\max(D^+,D^-).
\]
The gaps $G$ are $0.004$ for the uniform 70-ball, $0.017$ for MNIST digit 3,
$0.011$ for raw ImageNet koala, and $0.022$ for the GSM ($\sigma_S=0.25$). Together with the
small outside-range mass $\epsilon$, these checks support the separately
fitted pooled von Mises laws in the evaluated settings, including the GSM.
They do not assess the aggregated $(\hat\nu,\hat\tau)$ pair used in
calibration: for raw koala the pooled fit has $\tau=39.98$, while calibration
uses $\hat\tau=90.1$.
Figure~\ref{fig:vm-adequacy} shows representative uniform-ball and ImageNet
koala comparisons.

\section{Model and numerical validity}
\label{app:model-validity}

\subsection{Algorithm}
\label{app:algorithm}

Algorithm~\ref{alg:componentwise-danco} connects the componentwise objective
to a complete estimation procedure. Its steps also identify the operations
used in the complexity and timing analyses below. As an implementation convention, the procedure
applies one uniform lower cutoff: it skips angular calibration whenever the
distance estimate is at most five and returns the distance estimate alone.

\begin{algorithm}[H]
\caption{Componentwise calibrated intrinsic-dimension estimation}
\label{alg:componentwise-danco}
\begin{algorithmic}[1]
\Require data $\mathcal X$, neighborhood size $k$, candidates
$1{:}m_{\max}$, distance model, angular mode, reference source
\State Find the $k+1$ nearest neighbors and distances of every $x_i$
\State Form MiND or Gride ratios and fit $\dhat_{\mathrm{dist}}$
\If{$\dhat_{\mathrm{dist}}\leq5$}
  \State \Return $\dhat_{\mathrm{dist}}$
\EndIf
\State Compute the $\binom{k}{2}$ neighbor angles at every observation
$x_i$, the center of its neighborhood
\State Fit $(\hat\nu_i,\hat\tau_i)$ and aggregate
$(\hat\nu,\hat\tau)$ by circular and arithmetic means, respectively
\If{a circular resultant vector is zero}
  \State \Return $\dhat_{\mathrm{dist}}$
\EndIf
\For{$m=1,\ldots,m_{\max}$}
  \State Obtain dimension-$m$ reference statistics from a newly simulated
  uniform hyperball, reused Monte Carlo reference statistics, or the precomputed
  spline surface
  \State Compute $\Delta_{\mathrm{dist}}(m)$ and the Full and Profiled
  angular discrepancies
\EndFor
\State Minimize each component curve and each pointwise combined curve
\State Optionally refine an interior integer minimum by interpolation on
$[1,m_{\max}]$
\State \Return the selected estimate and all component curves
\end{algorithmic}
\end{algorithm}

Here a Monte Carlo reference is a simulation-based finite-sample calibration.
For each candidate $m$, it draws $N$ points uniformly from the
$m$-dimensional unit hyperball and computes
$(\hat d_m^{\mathrm{ref}},\nu_m^{\mathrm{ref}},
\tau_m^{\mathrm{ref}})$. Repeating this construction across all candidates
forms one reference replicate.

\subsection{Computational complexity}
\label{app:complexity}

\paragraph{Time complexity}
The analysis follows the steps of Algorithm~\ref{alg:componentwise-danco} and separates the observed-data steps 1--7 from the candidate loop, steps 11--16, for two reasons. Steps 1--7 form the observed-data pipeline of every variant: the distance model selects the fit at step 2, whereas the angular mode and the reference source change nothing before step 8; the exit at step 3 skips the angular steps 6--7, and the exit at step 8 skips the loop. They are also the only steps that involve the ambient dimension $D$, whereas the loop is governed by the candidate bound $m_{\max}$ and the reference source. The loop is analyzed for the three reference sources admitted at step 12, because the divergences at step 13 cost $O(k)$ per candidate independently of $N$, so the reference source is what sets the loop's cost.

\emph{Observed data, steps 1--7.} Under the exact pairwise-distance model used by the experiments, the neighbor search at step 1 costs $O(N^2D)$ and the $\binom{k}{2}$ angles per center at step 6 cost $O(Nk^2D)$. The remaining steps are cheaper. At step 2, forming the MiND first-neighbor ratios or the Gride generic-order ratios from the stored $(k+1)$-nearest-neighbor distances costs $O(Nk)$, and one evaluation of the distance log-likelihood during the fit of $\dhat_{\mathrm{dist}}$ costs $O(N)$; the integer grid search that initializes the MiND fit evaluates it at $d=1,\ldots,D$, for $O(ND)$. At step 7, fitting the per-center von Mises statistics accumulates the $\binom{k}{2}$ angle sines and cosines at every center for $O(Nk^2)$ and inverts one Bessel ratio per center for $O(N)$; the exits at steps 3 and 8 inspect quantities already computed. All of these terms are dominated by the neighbor search, so the observed-data time is
\[
 T_{\mathrm{data}}=O\!\left\{N(N+k^2)D\right\},
\]
in which $N^2D$ is the search and $Nk^2D$ the angle construction.

\emph{Candidate loop and selection, steps 11--16.} Each candidate divergence at step 13 costs $O(k)$ and does not depend on $N$: a finite digamma series of $k+1$ terms for MiND, a finite digamma sum of $k_2-k_1$ terms for Gride, or the adaptive one-dimensional quadrature of the released implementation (at most 50 subintervals). With the neighbor orders fixed, the minimization and optional refinement at steps 15--16 scan the $m_{\max}$ candidates once, for $O(m_{\max})$. The loop cost therefore depends on how step 12 obtains the reference statistics, and three sources are admitted. A newly simulated hyperball for every candidate is the calibration of the original DANCo~\cite{ceruti2014danco}, which rebuilds every reference for every dataset; it is the cost that the other two sources avoid. The precomputed spline surface is the FastDANCo route and serves the CNN profiles of Section~\ref{app:design-cnn}, where hundreds of layers are calibrated up to $m_{\max}=400$. Reused Monte Carlo references are the design of the synthetic, noise, GSM and image experiments of Section~\ref{app:design-common}: one reference replicate at the observed sample size is simulated once and shared by every component objective and configuration compared on it, which is what the growing calibration cache of the released implementation does. The last two sources are the modes run in this study; the first is analyzed as their baseline.

With a newly simulated hyperball, step 12 repeats steps 1, 2, 6 and 7 inside a simulated dimension-$m$ hyperball of $N$ points, so one reference costs $O\{N(N+k^2)m\}$. Because
$\sum_{m=1}^{m_{\max}}m=m_{\max}(m_{\max}+1)/2$, the candidate sweep
contributes a factor quadratic in the candidate bound, giving
\begin{equation}
 T_{\mathrm{MC}}
 =O\!\left\{N(N+k^2)(D+m_{\max}^2)\right\},
 \label{eq:supp-mc-complexity}
\end{equation}
whose first summand is the observed geometry and whose second the accumulated
reference simulation.
With the precomputed surface of FastDANCo, defined in Main Section 3.1, step 12 is one $O(1)$ lookup per candidate, so steps 11--16 together cost $O(m_{\max})$ and the online time becomes
\begin{equation}
 T_{\mathrm{Fast}}
 =O\!\left\{N(N+k^2)D+m_{\max}\right\},
 \label{eq:supp-fast-complexity}
\end{equation}
in which the simulation summand of~\eqref{eq:supp-mc-complexity} collapses to a
single scan over candidates. With reused Monte Carlo references, step 12 reads statistics simulated before the run, so the online cost has the form of $T_{\mathrm{Fast}}$ with the simulation term already paid. For a sample-size grid $\mathcal G_N$ and $n_{\mathrm{sim}}$ simulated
references per grid cell, the one-time construction cost of the surface is
\[
 T_{\mathrm{offline}}
 =O\!\left\{n_{\mathrm{sim}}\sum_{N_g\in\mathcal G_N}
 N_g(N_g+k^2)m_{\max}^2\right\},
\]
paid once for every sample size on the grid.
The packaged surface uses five sample sizes, $m_{\max}=400$, and $n_{\mathrm{sim}}=35$, for
$5\times400\times35=70{,}000$ calibration simulations. Ceruti et al. report
the original FastDANCo bound under their nearest-neighbor search model and a candidate
range tied to ambient dimension~\cite{ceruti2014danco}. Equations~\eqref{eq:supp-mc-complexity}--\eqref{eq:supp-fast-complexity} instead match
the exact search used here and keep ambient dimension $D$ separate from
candidate bound $m_{\max}$.

\paragraph{Space complexity}
For on-the-fly Monte Carlo calibration, an unchunked exact distance matrix uses $O(N^2)$ workspace; query batches of size $b$ reduce this term to $O(bN)$, giving the peak storage
\[
 O\!\left\{N\max(D,m_{\max})+bN+Nk^2\right\},
\]
whose three terms hold the data matrix together with the largest candidate
reference point set of step 12, the step-1 buffer that scores $b$ query rows against all $N$
points at once, and the $\binom{k}{2}$ neighbor-pair angles of step 6 at each of the $N$
centers.

For FastDANCo evaluation on the precomputed surface, the peak storage is
\[
 O\!\left(ND+bN+Nk^2+m_{\max}+S_{\mathrm{spline}}\right),
\]
which adds the candidate curves of steps 13--15 and the fitted surface $S_{\mathrm{spline}}$ read at step 12 to the same three data-side terms.

\subsection{Numerical implementation and safeguards}
\label{app:numerical-safeguards}

Two safeguards stabilize candidate selection. The angular discrepancy uses
exponentially scaled Bessel functions before any overflowing intermediate is
formed. Fractional refinement is restricted to the computed candidate interval
$[1,m_{\max}]$, preventing extrapolated minima.

When a circular resultant vector is zero, its mean direction is undefined;
the angular component is marked
unavailable and the estimator returns the distance estimate.

\begin{figure}[ht]
\centering
\includegraphics[width=0.72\linewidth]{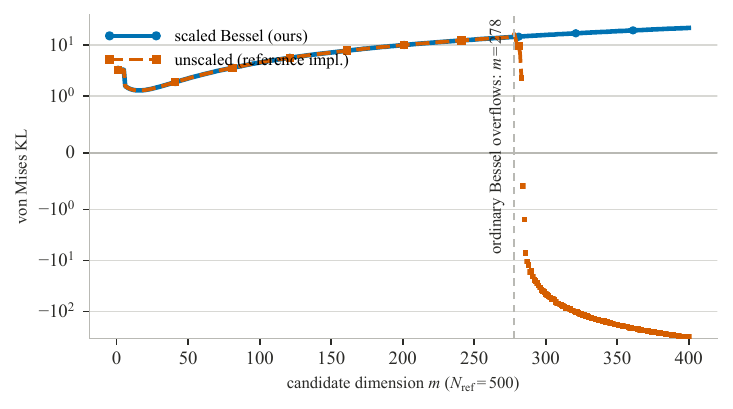}
\caption{Calibration-side Bessel overflow for fixed data statistics and the
$N_{\mathrm{ref}}=500$ calibration. Ordinary evaluation loses validity at
candidate $m=278$ and becomes negative at $m=284$. Scaled evaluation remains
finite and positive through candidate 400; dashed squares and solid circles
denote ordinary and scaled evaluation, respectively.}
\label{fig:overflow-supp}
\end{figure}

Define the exponentially scaled modified Bessel function by
\begin{equation}
  I_\alpha^e(x)=e^{-|x|}I_\alpha(x),
  \label{eq:supp-scaled-bessel}
\end{equation}
where $\alpha$ denotes the Bessel order and is unrelated to the angular mean
direction $\nu$.
Then
\begin{equation}
 \begin{aligned}
  \log\frac{I_0(\tau_2)}{I_0(\tau_1)}
  &=|\tau_2|-|\tau_1|
    +\log\frac{I_0^e(\tau_2)}{I_0^e(\tau_1)},\\
  \frac{I_1(\tau)}{I_0(\tau)}
  &=\frac{I_1^e(\tau)}{I_0^e(\tau)}.
 \end{aligned}
  \label{eq:supp-bessel-ratios}
\end{equation}
Both identities follow directly from
$I_\alpha(x)=e^{|x|}I_\alpha^e(x)$: taking logarithms turns the exponential
factors into $|\tau_2|-|\tau_1|$, while the common factor cancels in the
$I_1/I_0$ ratio.
Ordinary double-precision evaluation overflows near concentration $709.78$.
Let $N_{\mathrm{ref}}$ be the number of points in a reference. The illustrated
calibration uses $N_{\mathrm{ref}}=500$, and its first affected candidate is
$m=278$; the boundary is similar across the five precomputed sample sizes $450$--$700$. The corrupted ordinary calculation first becomes negative at $m=284$,
violating KL non-negativity and creating a possible false minimum.
Equations~\eqref{eq:supp-scaled-bessel} and
\eqref{eq:supp-bessel-ratios} keep the KL finite and positive through candidate
400 (Figure~\ref{fig:overflow-supp}).

\subsection{Computational scaling by workload}
\label{app:runtime}

The four timed workloads are step ranges of Algorithm~\ref{alg:componentwise-danco} with the costs of Section~\ref{app:complexity}. Observed-data statistics runs steps 1--7 and stops before the candidate loop, so its time is $T_{\mathrm{data}}$. The three complete evaluations run all steps and differ only in the reference source at step 12: precomputed spline references give $T_{\mathrm{Fast}}$ of Eq.~\eqref{eq:supp-fast-complexity}, newly generated Monte Carlo references give $T_{\mathrm{MC}}$ of Eq.~\eqref{eq:supp-mc-complexity}, and reused Monte Carlo references give the form of $T_{\mathrm{Fast}}$ with the simulation term paid before timing.

% Reported medians use three fresh processes after one warm-up. JIT compilation
% and reused-reference cache population occur before timing; CPU and GPU inputs
% share float64 data and seeds, and Monte Carlo references are constructed on the
% CPU.

\begin{figure}[ht]
\centering
\includegraphics[width=0.98\linewidth]{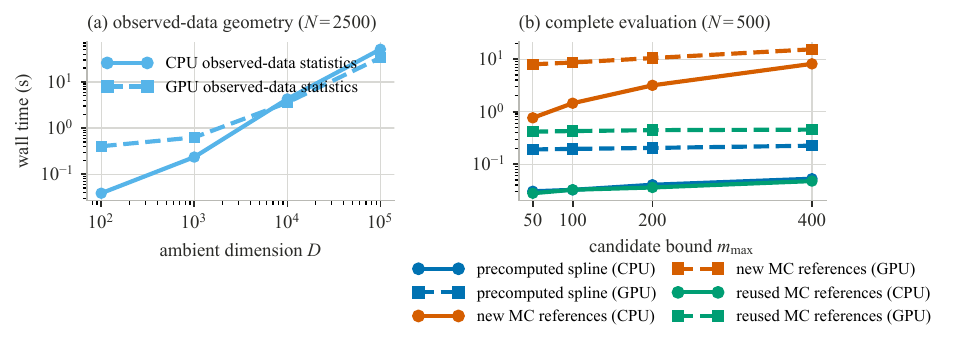}
\caption{Wall time for the explicitly defined workloads, shown as the median
of three fresh-process measurements after warm-up. (a) Observed-data statistics versus ambient
dimension. (b) Complete evaluation with precomputed spline, newly generated
Monte Carlo, or reused Monte Carlo references versus candidate bound; solid
lines with circles are CPU, while dashed lines with squares are GPU.}
\label{fig:runtime}
\end{figure}

Figure~\ref{fig:runtime}a varies ambient dimension at $N=2500$, $k=10$ while
timing only observed-data steps, thereby isolating the cost that grows with
$D$. CPU is faster for the two smallest workloads, where device transfer and process initialization dominate, while GPU becomes faster at $D\geq10^4$; at $D=10^5$ the observed-data computation takes about $51~\mathrm{s}$ on CPU and $34~\mathrm{s}$ on GPU, with median per-process peak resident memory of $3761~\mathrm{MB}$ on CPU and $3824~\mathrm{MB}$ of allocated device memory on GPU.

Figure~\ref{fig:runtime}b fixes $N=500$, $D=400$, and $k=10$ and varies
$m_{\max}$ to isolate reference construction and candidate evaluation. At $m_{\max}=400$, precomputed and reused references remain below $0.5~\mathrm{s}$ on both backends, whereas regenerating Monte Carlo references takes seconds and grows with the candidate bound because it constructs more references; the observed data are small in this setting, so GPU setup costs exceed the observed-data savings.

The precomputed and reused evaluations in Figure~\ref{fig:runtime}b therefore finish one dataset in under $0.5$ s on both backends; Main Table 1 separately reports a median of $0.03$ seconds for MiND--Full, which excludes the shared reference construction. These measurements place componentwise
calibration in the practical cost range of the estimators collected in
\texttt{scikit-dimension}~\cite{bac2021scikit} for the evaluated workloads and
data sizes.

\section{Experimental design and reporting}
\label{app:experimental-design}

\subsection{Common settings}
\label{app:design-common}

Where an implementation is available, the reference estimators come from
\texttt{scikit-dimension}~\cite{bac2021scikit}.
The synthetic, noise, GSM, and image experiments construct Monte Carlo
references at the observed sample size by the construction of
Section~\ref{app:algorithm}. One reference replicate therefore holds
$(\hat d_m^{\mathrm{ref}},\nu_m^{\mathrm{ref}},\tau_m^{\mathrm{ref}})$ at
every candidate, and the component objectives entering a comparison reuse it.
Each data replicate draws every dataset afresh and independently. Data
and reference replicates are shared across configurations whenever a comparison
changes one estimator component or one perturbation level, which keeps the data
and calibration realization matched for the comparison. Unless stated
otherwise, MPE is computed within each replicate and reported as a mean $\pm$
standard deviation over replicates. Each experiment is summarized within its
own replicate design, so summaries at nominally identical settings differ
slightly across experiments: the clean MiND--Full MPE is $6.33\%$ under the
twenty-replicate aggregation of Main Table 1, $6.24\%$ at $k=10$, $N=2500$ in
the five-replicate sensitivity sweep (Table~\ref{tab:sensitivity}), and
$6.75\%$ at $\eta=0$ in the ten-replicate noise design
(Table~\ref{tab:relative-noise-supp}).

The candidate search runs over $m=1,\ldots,\min(m_{\max},D)$ with
$m_{\max}=100$ for the standard synthetic experiments, $200$ for the image
experiments, and $400$ for the CNN profiles; benchmark manifolds whose ambient dimension lies below
$m_{\max}$ are therefore searched up to $D$. Reference angular statistics and penalty curves are reported over the valid candidates, those whose reference distance estimate exceeds five, which start at $m=6$. The search itself runs over every candidate; references with a distance estimate of at most five carry zero concentration and are excluded only from the reported ranges.

\subsection{Clean-manifold benchmark}
\label{app:design-clean}

The primary benchmark uses 24 manifolds with $N=2500$ points each. Twenty data
replicates form the sampling design, and the ten estimators of the primary
comparison evaluate the same replicates, so estimator differences carry no
sampling offset; MiND--Full uses the matching Monte Carlo reference replicate.

The component comparison keeps the 14 benchmark manifolds on which all eight
component objectives are defined: on the other ten, a preliminary MiND
distance estimate at most five omits angular calibration, so the angle-only
objectives are undefined. Five reference replicates, with one data replicate per manifold, supply the replicate summary.

\subsection{Sensitivity sweeps}
\label{app:design-sensitivity}

The sensitivity study sweeps the neighborhood size over $k\in\{5,10,15,20\}$ at
$N=2500$ and the sample size over $N\in\{625,1250,2500,5000\}$ at $k=10$. Each
setting uses five data replicates of the 24-manifold suite.

\subsection{Neighborhood-relative noise}
\label{app:design-noise}

The noise levels are $\eta\in\{0,0.05,0.1,0.2,0.4\}$ under Main Eq.~(11). Ten data replicates of
2500 observations per manifold are generated. Within a replicate, the clean data
are shared across noise levels, so a change in accuracy follows the added noise.

\subsection{Gaussian scale mixture}
\label{app:design-gsm}

The amplitude experiment uses the lognormal GSM of Section~\ref{app:nu-proofs} with generating dimension $d=70$ and $N=2500$ observations per sample. Each sample is zero-padded into $D=100$, an isometric embedding that leaves every neighbor distance and angle unchanged and extends the candidate search to $100$. The sweep varies $\sigma_S\in\{0,0.1,0.15,0.2,0.25,0.3,0.35\}$ over 30 data replicates at each level. Within a replicate, the latent Gaussian draws are reused across amplitude levels, so every comparison changes one factor at a time.

To relate the simulated amplitude sweep to the amplitude variation observed in real images, we summarize sample-to-sample radial variation by the coefficient of variation (CV) of the centered norm. For centered norms $a_i=\|x_i-\bar x\|_2$, the CV, computed with population standard deviation, is
\[
 \bar a=\frac1N\sum_{i=1}^N a_i,\qquad
 \operatorname{CV}(a)=
 \frac{\{N^{-1}\sum_{i=1}^N(a_i-\bar a)^2\}^{1/2}}{\bar a},
 \quad \bar a>0.
\]
For the lognormal amplitude, $\mathrm{CV}_S=\{\mathrm{Var}(S_i)\}^{1/2}/\mathbb E(S_i)=(e^{\sigma_S^2}-1)^{1/2}$, and $\mathrm{CV}_Z$ denotes the coefficient of variation of $\|Z_i\|$, equal to $0.085$ at $d=70$. Under population centering, $\|X_i\|=S_i\|Z_i\|$, and the independence of $S_i$ and $Z_i$ gives
\[
 1+\operatorname{CV}^2=(1+\mathrm{CV}_S^2)(1+\mathrm{CV}_Z^2).
\]
The sample-centered $\operatorname{CV}(a)$ follows this relation approximately, so the sweep brackets the values $\operatorname{CV}(a)=0.15$--$0.33$ observed across the 17 image classes of Table~\ref{tab:class-survey}. The level $\sigma_S=0.25$ lies inside this range and serves as the focal level for the detailed comparison and the control.

At the focal level, two diagnostics complement the sweep. The first follows the angular-location penalty along the candidate axis over five replicates and is reported in Section~\ref{app:gsm-results}. The second is a known-amplitude control that divides each observation by its simulated amplitude, $X_i/S_i=Z_i$. All data-side statistics are recomputed from $Z_i$: the neighbor search, the distance ratios, and the neighbor angles. The latent Gaussian draws, the generating dimension $d=70$, and the paired reference realization are retained, so the transformation removes only the multiplicative amplitude factor. Because the same $Z_i$ draws are used across amplitude levels, the control coincides with the matched $\sigma_S=0$ sample by construction.

A separate simulation examines the regime transition of Main Figure~2 with $N=2500$, $k=10$, exact nearest-neighbor search, and the same angular estimator. The homogeneous-amplitude panel varies the ball dimension over $d\in\{2,3,5,8,12,20,35,50,70,100,150,250,400\}$, and the heterogeneous-amplitude panel fixes $d=70$ and varies $\sigma_S$ over the sweep grid above. The center-level diagnostic of Figure~\ref{fig:nu-mechanism-supp} uses one sample at the focal level $\sigma_S=0.25$.

\subsection{Image data}
\label{app:design-images}

The image classes are MNIST digits 3 and 7~\cite{lecun1998gradient}, CIFAR-10
birds and cats~\cite{krizhevsky2009learning}, and the ImageNet koala and
milkweed-butterfly classes~\cite{russakovsky2015imagenet,ansuini2019intrinsic}. They contain $1010/1028$,
$5000$, and $547/648$ observations, and their ambient dimensions are $784$,
$3072$, and $150528$. Main Section 4.1 defines the two normalizations, centered
radial and per-image contrast. A zero centered vector remains zero, a constant
image maps to zero, and every observed-data statistic (neighbor search,
distance ratios, angles) is recomputed on the transformed sample.

Each class contributes one fixed sample, so the observed angular statistics
stay fixed while the calibration side varies. Five reference replicates serve every image estimate, and each calibrated estimate averages over the five. Over the valid candidates $m=6,\ldots,200$, the five replicates give mean-direction ranges within $[1.142,1.495]$ radians; Table~\ref{tab:reference-range-supp} reports the range matched to each sample size together with the observed data value.

\begin{table}[H]
\centering
\caption{Sample-size-specific reference mean-direction ranges. Each image row uses
the five reference replicates over candidates $m=6,\ldots,200$; the last row gives the Gride reference surface of the CNN study at its effective sample sizes over $m=6,\ldots,400$. The observed
gap is the lower reference boundary minus the data mean direction. A negative gap places the data value above the lower boundary, and both MNIST values also lie below the upper boundary, hence inside the range.}
\label{tab:reference-range-supp}
\small
\begin{tabular}{r l rrrr}
\toprule
$N$ & Data & reference min & reference max & observed $\hat\nu$ & gap \\
\midrule
547 & ImageNet koala & 1.142 & 1.443 & 0.934 & 0.208 \\
648 & ImageNet butterfly & 1.144 & 1.447 & 0.915 & 0.229 \\
1010 & MNIST 3 & 1.150 & 1.464 & 1.166 & $-0.016$ \\
1028 & MNIST 7 & 1.150 & 1.467 & 1.157 & $-0.006$ \\
5000 & CIFAR-10 bird & 1.169 & 1.495 & 0.898 & 0.271 \\
5000 & CIFAR-10 cat & 1.169 & 1.495 & 0.946 & 0.223 \\
\midrule
495--500 & CNN surface (Gride) & 1.115 & 1.431 & -- & --%
\\
\bottomrule
\end{tabular}
\end{table}

\subsection{CNN study}
\label{app:design-cnn}

The CNN study evaluates AlexNet~\cite{krizhevsky2012imagenet},
VGG-16~\cite{simonyan2015very}, and ResNet-18/34~\cite{he2016deep} on seven
single-object ImageNet categories, with three independent 500-image subsamples
for every architecture and category. At each checkpoint, exact-duplicate
activation rows are removed once, and
Gride--Profiled, TWO-NN, and MLE read the resulting common matrix. The effective
sample size used by the Gride reference surface is therefore the number of
unique rows. Over the effective sample sizes of this study ($495$--$500$
unique rows) and candidates $m=6,\ldots,400$, the surface's reference mean
direction ranges from $1.115$ to $1.431$ radians (Table~\ref{tab:reference-range-supp}) and decreases overall with
the candidate dimension. A single Gride--Profiled cell (ResNet-18, second residual stage, one subsample) returns the candidate bound $400$; the reported medians are unaffected. Both Gride variants use integer candidate minima in this study.
TWO-NN uses the discarded-tail fit of Facco et al.\ via
\texttt{scikit-\allowbreak dimension}; MLE uses the package's $k=10$ Levina--Bickel estimate.
Checkpoints record pooling, stage, and linear outputs, so the
sampled grid resolves layer-wise change to the nearest such transition.
Relative depth divides the cumulative number of convolutional, linear, and
final adaptive-pooling transformations at a checkpoint by the maximum such
count, placing every network on a common input-to-output interval; residual
projection shortcuts are not counted separately.

The packaged Gride reference surface of Main Section 3.5 covers
$m=1,\ldots,400$ at sample sizes 450, 500, 580, 640, and 700, and every
grid point averages 35 independently simulated hyperballs. The MiND surface comes from the same grid and the same simulated hyperballs, so the two surfaces agree grid point by grid point on their angular entries and differ only in the distance statistic.

\section{Extended experimental results}
\label{app:extended-results}

\subsection{Benchmark detail and reference estimators}
\label{app:benchmark-detail}

\begin{figure}[ht]
\centering
\includegraphics[width=0.82\linewidth]{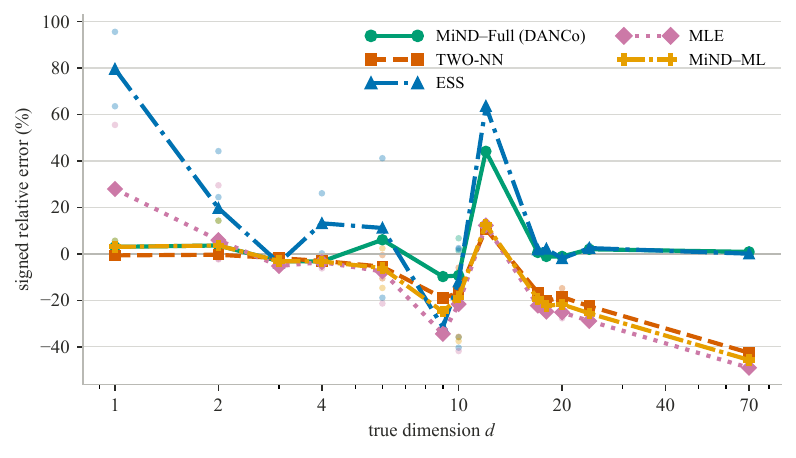}
\caption{Signed relative error on the 24-manifold benchmark. Faint points are
individual manifolds; lines are means by generating dimension. Color, line
style, and marker jointly identify each estimator.}
\label{fig:error-vs-d-supp}
\end{figure}

Table~\ref{tab:baselines-full} and Figure~\ref{fig:error-vs-d-supp} expand Main Table 1 with per-manifold estimates and signed errors.

\begin{landscape}
\vspace*{\fill}
\begin{table}[H]
\centering
\caption{Estimated ID of ten estimators on all 24 benchmark manifolds (mean
over the successful data replicates out of 20). FisherS on M10d\_Cubic
averages its single successful replicate; its other 19 replicates failed.}
\label{tab:baselines-full}
\small
\setlength{\tabcolsep}{3.5pt}
\begin{tabular}{l r rrrrrrrrrr}
\toprule
Manifold & $d$ & CorrInt & MiND--Full (DANCo) & ESS & FisherS & MADA & MLE & MiND--ML & TLE & TWO-NN & lPCA \\
\midrule
M10a\_Cubic & 10 & 8.6 & 10.1 & 10.2 & 10.3 & 9.2 & 8.7 & 8.9 & 9.6 & 9.1 & 11.0 \\
M10b\_Cubic & 17 & 12.7 & 17.1 & 17.3 & 17.0 & 13.9 & 13.2 & 13.7 & 14.2 & 14.2 & 18.0 \\
M10c\_Cubic & 24 & 16.2 & 24.5 & 24.4 & 24.0 & 18.0 & 17.2 & 18.0 & 18.1 & 18.8 & 25.0 \\
M10d\_Cubic & 70 & 31.8 & 70.6 & 70.0 & 57.1 & 36.8 & 35.7 & 38.0 & 35.0 & 40.2 & 71.0 \\
M11\_Moebius & 2 & 2.0 & 2.0 & 2.5 & 2.0 & 2.1 & 2.0 & 2.0 & 2.2 & 2.0 & 3.0 \\
M12\_Norm & 20 & 12.7 & 20.0 & 19.8 & 20.0 & 16.2 & 15.5 & 16.3 & 15.5 & 17.0 & 20.0 \\
M13a\_Scurve & 2 & 2.0 & 2.0 & 2.1 & 2.9 & 2.1 & 2.0 & 2.0 & 2.1 & 2.0 & 3.0 \\
M13b\_Spiral & 1 & 4.6 & 1.1 & 2.0 & 2.0 & 3.8 & 1.6 & 1.1 & 1.9 & 1.0 & 2.0 \\
M1\_Sphere & 10 & 8.9 & 10.7 & 10.2 & 11.0 & 9.6 & 9.0 & 9.2 & 10.1 & 9.4 & 11.0 \\
M2\_Affine\_3to5 & 3 & 2.9 & 2.9 & 2.9 & 2.9 & 3.0 & 2.8 & 2.9 & 3.1 & 2.9 & 3.0 \\
M3\_Nonlinear\_4to6 & 4 & 3.5 & 3.8 & 4.0 & 3.4 & 4.1 & 3.7 & 3.8 & 4.2 & 3.8 & 5.0 \\
M4\_Nonlinear & 4 & 3.8 & 3.9 & 5.0 & 5.8 & 4.4 & 4.0 & 3.9 & 4.5 & 3.9 & 8.0 \\
M5a\_Helix1d & 1 & 1.0 & 1.0 & 1.6 & 2.9 & 1.1 & 1.0 & 1.0 & 1.1 & 1.0 & 3.0 \\
M5b\_Helix2d & 2 & 2.8 & 2.3 & 2.9 & 2.9 & 3.1 & 2.6 & 2.3 & 2.9 & 2.0 & 3.0 \\
M6\_Nonlinear & 6 & 5.9 & 6.4 & 8.5 & 8.5 & 7.2 & 6.4 & 6.1 & 7.1 & 6.0 & 12.0 \\
M7\_Roll & 2 & 1.9 & 2.0 & 2.1 & 2.9 & 2.1 & 2.0 & 2.0 & 2.1 & 2.0 & 3.0 \\
M8\_Nonlinear & 12 & 11.4 & 17.3 & 19.6 & 17.4 & 14.6 & 13.5 & 13.5 & 14.0 & 13.3 & 24.0 \\
M9\_Affine & 20 & 13.7 & 19.6 & 19.4 & 18.9 & 15.2 & 14.5 & 15.0 & 15.3 & 15.5 & 20.0 \\
Mbeta & 10 & 3.4 & 6.4 & 6.0 & 5.3 & 6.6 & 5.8 & 6.2 & 6.2 & 6.4 & 10.0 \\
Mn1\_Nonlinear & 18 & 12.5 & 17.8 & 18.4 & 17.1 & 14.2 & 13.5 & 14.0 & 14.2 & 14.3 & 27.0 \\
Mn2\_Nonlinear & 24 & 15.4 & 24.4 & 24.8 & 23.1 & 17.7 & 16.9 & 17.7 & 17.5 & 18.4 & 36.0 \\
Mp1\_Paraboloid & 3 & 2.1 & 2.9 & 2.9 & 0.9 & 3.1 & 2.8 & 2.9 & 3.1 & 3.0 & 1.0 \\
Mp2\_Paraboloid & 6 & 2.7 & 6.3 & 4.9 & 0.9 & 5.1 & 4.7 & 5.1 & 5.3 & 5.4 & 1.0 \\
Mp3\_Paraboloid & 9 & 3.1 & 8.1 & 6.2 & 0.9 & 6.5 & 5.9 & 6.8 & 6.7 & 7.3 & 1.0 \\
\bottomrule
\end{tabular}

\end{table}
\vspace*{\fill}
\end{landscape}

\FloatBarrier

\subsection{Clean component agreement}
\label{app:component-results}

Table~\ref{tab:clean-components-supp} reports MPE and signed error for all eight objectives on the complete-case set of Main Section 4.2, the 14 manifolds whose preliminary MiND distance estimate exceeds five (Section~\ref{app:design-clean}).

\begin{table}[H]
\centering
\caption{Clean component comparison on 14 eligible manifolds over five shared
reference replicates. Entries are mean $\pm$ standard deviation across
replicates. MPE gives error magnitude; signed error retains direction, with
negative values denoting underestimation. All eight objectives minimize a
calibrated KL objective with fractional refinement.}
\label{tab:clean-components-supp}
\small
\begin{tabular}{lrr}
\toprule
Objective & MPE (\%) & Signed error (\%) \\
\midrule
MiND distance-only & $10.1\pm0.2$ & $-1.6\pm0.3$ \\
Gride distance-only & $11.9\pm0.6$ & $-1.8\pm0.4$ \\
Full angle-only & $15.9\pm0.7$ & $+10.7\pm0.7$ \\
Profiled angle-only & $12.2\pm0.3$ & $+6.2\pm0.2$ \\
MiND--Full & $8.6\pm0.7$ & $+0.9\pm0.9$ \\
MiND--Profiled & $9.7\pm0.2$ & $-0.6\pm0.2$ \\
Gride--Full & $11.6\pm0.7$ & $-1.2\pm0.8$ \\
Gride--Profiled & $11.7\pm0.4$ & $-2.5\pm0.3$ \\
\bottomrule
\end{tabular}
\end{table}

All four combined objectives have lower observed mean MPE than either of
their own components on matched clean geometry; for the two Gride objectives the margin over Gride distance-only is $0.2$--$0.3$ MPE points, within one reference-replicate standard deviation.  Both
distance-only objectives sit below the generating dimension and both
angle-only objectives above it, matching the directions reported for the raw
statistics in the original DANCo study~\cite{ceruti2014danco}. The four
combined objectives have mean signed errors within $2.6$ percentage points of
zero, each closer to the corresponding distance-only mean signed error than to
the angle-only one.

To interpret this pattern, note that equal objective weights need not imply equal local influence. Averages across manifolds should not be interpreted as convex combinations of the component averages; for an individual manifold whose component minima are sufficiently close, a local quadratic approximation makes this influence explicit. Along the candidate coordinate $x=\log m$, write the combined objective as $\Delta_{\mathrm{dist}}(x)+\Delta_{\mathrm{ang}}(x)$, let $x_{\mathrm{dist}}$ and $x_{\mathrm{ang}}$ be the minimizers of the two components, and let $x_{\mathrm{joint}}$ be the minimizer of their sum, so that $\exp(x_{\mathrm{joint}})$ is the estimate returned by the combined objective. Expand each component about its own minimum,
\[
 \Delta_{\mathrm{dist}}(x)\approx\Delta_{\mathrm{dist}}(x_{\mathrm{dist}})+\tfrac12H_{\mathrm{dist}}(x-x_{\mathrm{dist}})^{2},\qquad
 \Delta_{\mathrm{ang}}(x)\approx\Delta_{\mathrm{ang}}(x_{\mathrm{ang}})+\tfrac12H_{\mathrm{ang}}(x-x_{\mathrm{ang}})^{2},
\]
where $H_{\mathrm{dist}}=\Delta_{\mathrm{dist}}''(x_{\mathrm{dist}})$ and $H_{\mathrm{ang}}=\Delta_{\mathrm{ang}}''(x_{\mathrm{ang}})$. Setting the derivative of the sum of the two quadratic approximations to zero gives
\[
 x_{\mathrm{joint}}\approx\frac{H_{\mathrm{dist}}x_{\mathrm{dist}}+H_{\mathrm{ang}}x_{\mathrm{ang}}}{H_{\mathrm{dist}}+H_{\mathrm{ang}}}.
\]
Thus, locally and when the two minima are nearby, the more strongly curved component has the greater influence on the joint minimum. To compare the component curvatures analytically, consider an exact match between an observed statistic and its reference value. There the divergence and its gradient vanish, so its second derivative in the reference parameter equals the Fisher information of the fitted law and the chain rule leaves no term from the curvature of the reference trajectory; the curvature of each component in $x$ is therefore the Fisher information of its parameter times the squared rate at which the reference moves that parameter. The next proposition gives the corresponding Fisher informations.

\begin{proposition}[Fisher curvature of the distance and concentration parameters]
\label{prop:supp-log-fisher}
(a) Under the local Poisson model of Section~\ref{app:gride-kl}, let $W=d\log(r_{k_2}/r_{k_1})$ for neighbor orders $k_1<k_2$; at $(k_1,k_2)=(1,k+1)$ this is the MiND quantity $-d\log\rho$. Then $e^{-W}\sim\mathrm{Beta}(k_1,k_2-k_1)$ for every $d$, and the Fisher information of $\log d$ carried by the observed ratio is
\begin{equation}
 \mathcal F=1+(k_2-k_1-1)\,\mathbb E\!\left[\frac{W^{2}e^{W}}{(e^{W}-1)^{2}}\right]\geq1,
 \label{eq:fisher-ratio}
\end{equation}
with equality exactly for adjacent orders. (b) Under a von Mises law with concentration $\tau>0$, the Fisher information of $\log\tau$ is $C(\tau)=\tau^{2}A'(\tau)>0$, which extends continuously to $C(0)=0$ and satisfies $C(\tau)\to1/2$ as $\tau\to\infty$.
\end{proposition}

\begin{proof}
(a) Substituting $v=\mu^{-d}$ in the Gride density of Main Eq.~(2) gives $v^{k_1-1}(1-v)^{k_2-k_1-1}/B(k_2-k_1,k_1)$ on $(0,1)$, the stated Beta law, which at $(1,k+1)$ is the MiND law $\rho^{d}\sim\mathrm{Beta}(1,k)$. The observed log ratio $W/d$ therefore has log density $\log d-k_1W+(k_2-k_1-1)\log(1-e^{-W})$ up to a constant, with score $1-k_1W+(k_2-k_1-1)W/(e^{W}-1)$ in $\log d$. Differentiating once more and using the zero mean of the score gives Eq.~\eqref{eq:fisher-ratio} as the expected negative second derivative. The expectation in Eq.~\eqref{eq:fisher-ratio} is positive because $W>0$ almost surely, so $\mathcal F=1$ for adjacent orders and $\mathcal F>1$ otherwise.

(b) For $\tau>0$, with $\theta$ measured from the mean direction, the von Mises log density is $\tau\cos\theta-\log\{2\pi I_0(\tau)\}$, so the score of $\log\tau$ is $\tau\{\cos\theta-A(\tau)\}$ and $C(\tau)=\tau^{2}\operatorname{Var}(\cos\theta)=\tau^{2}A'(\tau)>0$. The relations $I_0'=I_1$ and $I_1'=I_0-I_1/\tau$~\cite[Sec.~10.29]{olver2010nist} give $A'=1-A/\tau-A^{2}$, so with $y=\tau A$, $C=\tau^{2}-y-y^{2}$, and $C\to0$ as $\tau\to0$. The large-argument expansions of $I_0$ and $I_1$~\cite[Eq.~10.40.1]{olver2010nist} give $A(\tau)=1-1/(2\tau)-1/(8\tau^{2})+O(\tau^{-3})$, so $y=\tau-1/2-1/(8\tau)+O(\tau^{-2})$ and $C(\tau)=1/2+O(\tau^{-1})$.
\end{proof}

This comparison becomes especially simple for Profiled calibration. For a smooth reference trajectory, let $\beta_d=d\log\hat d_m^{\mathrm{ref}}/dx$ and $\beta_\tau=d\log\tau_m^{\mathrm{ref}}/dx$ be the local log-slopes at which the reference moves the two parameters along $x=\log m$, with $\beta_d\neq0$. At an exact component match, Proposition~\ref{prop:supp-log-fisher} gives the distance and Profiled angular curvatures
\[
 H_{\mathrm{dist}}=\mathcal F\beta_d^{2},\qquad H_{\mathrm{prof}}=C(\hat\tau)\beta_\tau^{2},
\]
the latter, the $H_{\mathrm{ang}}$ of the Profiled objective, because the Profiled angular term is a KL divergence between von Mises laws with aligned mean directions (Section~\ref{app:profiled-divergence}). Their relative local influence is therefore summarized directly by
\[
 R=\frac{H_{\mathrm{prof}}}{H_{\mathrm{dist}}}=\frac{C(\hat\tau)}{\mathcal F}\left(\frac{\beta_\tau}{\beta_d}\right)^{2}.
\]
At exact matching, $\hat\tau=\tau_m^{\mathrm{ref}}$, so the candidate-wise diagnostic evaluates $C$ at $\tau_m^{\mathrm{ref}}$. On the sampled reference curves, we estimate these local slopes as follows. The curves are the Monte Carlo references behind Table~\ref{tab:clean-components-supp}, rebuilt from the benchmark's five calibration seeds at $N=2500$ over $m=6,\ldots,100$ and averaged over the replicates. At each sampled candidate $m$ we estimate the two local log-slopes $\beta_{d,m}$ and $\beta_{\tau,m}$ by separate regressions of $\log\hat d_m^{\mathrm{ref}}$ and $\log\tau_m^{\mathrm{ref}}$ on $\log m$ over the $m\pm10$ window, using the available candidates near the boundaries, and evaluate
\[
 R_m=\frac{C(\tau_m^{\mathrm{ref}})}{\mathcal F}\left(\frac{\beta_{\tau,m}}{\beta_{d,m}}\right)^{2},
\]
where $\mathcal F$ comes from quadrature of Eq.~\eqref{eq:fisher-ratio} and equals $5.86$ for MiND at $k=10$ and $4.79$ for Gride at $(k_1,k_2)=(5,10)$. On the clean-benchmark references used for Table~\ref{tab:clean-components-supp}, the maximum $R_m$ is $0.31$ for MiND and $0.38$ for Gride; both remain below one under the wider $m\pm20$ window. These ratios are diagnostics on the sampled reference curves, not bounds on an underlying continuous derivative. Thus the distance component is locally more curved than the Profiled concentration component on these sampled references.

At a match of both angular parameters, Full calibration adds the mean-direction curvature $A(\hat\tau)\hat\tau\,(d\nu_m^{\mathrm{ref}}/dx)^{2}$ to $H_{\mathrm{prof}}$. Because the Profiled comparison does not control this additional term, its curvature ordering does not automatically extend to Full. This distinction is itself informative: Profiled isolates the concentration contribution, whereas Full retains the additional location geometry. The proximity of the Full estimates to the distance-only values in Table~\ref{tab:clean-components-supp} is therefore treated as an empirical pattern, whereas the sampled-reference diagnostics support stronger local distance influence for the Profiled combinations.

\FloatBarrier

\subsection{Sensitivity to neighborhood and sample size}
\label{app:sensitivity}

Table~\ref{tab:sensitivity} reports the Main Section 4.2 neighborhood and sample-size sweeps for five methods.

\begin{table}[H]
\centering
\caption{MPE (\%) across neighborhood and sample-size sweeps, each averaged
over five data replicates. The first block varies $k$ at $N=2500$; the second
varies $N$ at $k=10$.}
\label{tab:sensitivity}
\small
\begin{tabular}{r rrrrr}
\toprule
Setting & MiND--Full & Gride--Full & MiND--Profiled & Gride--Profiled & TWO-NN \\
\midrule
\multicolumn{6}{l}{Neighborhood sweep ($N=2500$)} \\
5 & 6.04 & 7.37 & 5.84 & 6.45 & 11.91 \\
10 & 6.24 & 9.50 & 6.73 & 9.39 & 11.91 \\
15 & 7.18 & 13.46 & 7.60 & 13.53 & 11.91 \\
20 & 8.09 & 16.78 & 8.49 & 16.91 & 11.91 \\
\midrule
\multicolumn{6}{l}{Sample-size sweep ($k=10$)} \\
625 & 14.06 & 17.22 & 13.73 & 17.03 & 15.46 \\
1250 & 9.03 & 15.25 & 8.99 & 15.12 & 13.28 \\
2500 & 6.24 & 9.50 & 6.73 & 9.39 & 11.91 \\
5000 & 5.38 & 6.63 & 5.56 & 6.97 & 10.76 \\
\bottomrule
\end{tabular}

\end{table}

The two Gride variants track each other across both sweeps: Gride--Profiled is
lower at $k=5,10$ and $N=625,1250,2500$, while Gride--Full is lower at
$k=15,20$ and $N=5000$; their largest difference is $0.92$ MPE points.
TWO-NN uses only the first two neighbor distances, so its column is constant
across $k$ by construction.

\FloatBarrier

\subsection{Neighborhood-relative noise}
\label{app:relative-noise-results}

The noise sweep identifies the crossover from MiND's clean-data advantage
to the benefit of larger-order Gride ratios.
Table~\ref{tab:relative-noise-supp} reports all five noise levels.

\begin{table}[H]
\centering
\caption{Neighborhood-relative Gaussian noise. Upper panel: MPE (\%) of the four combined objectives over all 24 manifolds, with a final row at $\eta=0.4$ restricted to the 17 manifolds that remain after excluding those with $d>5$ and $D\leq d+1$. Lower panel: signed error (\%) of all eight objectives on the 14-manifold complete-case set of Section~\ref{app:component-results}, with negative values denoting underestimation. Entries are mean $\pm$ standard deviation over the ten data replicates of the noise design; its $\eta=0$ column therefore differs slightly from the five-replicate clean values of Table~\ref{tab:clean-components-supp}.}
\label{tab:relative-noise-supp}
\small
\begin{tabular}{lrrrr}
\toprule
$\eta$ & MiND--Full & Gride--Full & MiND--Profiled & Gride--Profiled \\
\midrule
0 & 6.75 $\pm$ 0.41 & 9.39 $\pm$ 0.38 & 6.82 $\pm$ 0.38 & 9.41 $\pm$ 0.26 \\
0.05 & 7.67 $\pm$ 0.49 & 9.37 $\pm$ 0.34 & 7.69 $\pm$ 0.28 & 9.32 $\pm$ 0.28 \\
0.1 & 11.38 $\pm$ 0.93 & 9.16 $\pm$ 0.53 & 11.36 $\pm$ 0.69 & 9.05 $\pm$ 0.50 \\
0.2 & 16.48 $\pm$ 0.51 & 10.96 $\pm$ 0.55 & 15.84 $\pm$ 0.32 & 10.33 $\pm$ 0.65 \\
0.4 & 27.74 $\pm$ 0.77 & 17.64 $\pm$ 0.58 & 27.32 $\pm$ 0.94 & 16.96 $\pm$ 0.43 \\
\midrule
\multicolumn{5}{l}{Restricted subset: 17 manifolds (excluding $d>5$ with $D\leq d+1$)} \\
0.4 & 37.62 $\pm$ 1.00 & 23.43 $\pm$ 0.66 & 36.93 $\pm$ 1.20 & 22.49 $\pm$ 0.40 \\
\bottomrule
\end{tabular}
\\[6pt]
\begin{tabular}{lrrrrr}
\toprule
Objective & $\eta=0$ & $\eta=0.05$ & $\eta=0.1$ & $\eta=0.2$ & $\eta=0.4$ \\
\midrule
MiND distance-only & $-1.0\pm0.7$ & $-0.4\pm0.7$ & $+0.4\pm0.5$ & $+2.1\pm0.6$ & $+12.0\pm0.8$ \\
Gride distance-only & $-2.1\pm0.3$ & $-1.9\pm0.6$ & $-1.3\pm0.7$ & $+1.4\pm0.3$ & $+8.7\pm0.8$ \\
Full angle-only & $+10.9\pm1.1$ & $+11.4\pm0.7$ & $+11.8\pm0.9$ & $+17.0\pm1.9$ & $+32.9\pm1.2$ \\
Profiled angle-only & $+6.2\pm0.6$ & $+6.8\pm0.6$ & $+8.7\pm0.5$ & $+12.3\pm1.1$ & $+24.1\pm0.4$ \\
MiND--Full & $+1.4\pm1.1$ & $+1.6\pm1.1$ & $+2.2\pm0.6$ & $+4.6\pm1.4$ & $+14.7\pm1.2$ \\
MiND--Profiled & $+0.1\pm0.8$ & $+0.6\pm0.5$ & $+0.8\pm0.5$ & $+3.0\pm0.6$ & $+13.0\pm1.6$ \\
Gride--Full & $-1.1\pm0.8$ & $-0.5\pm1.2$ & $+1.1\pm0.8$ & $+5.7\pm0.9$ & $+12.8\pm0.8$ \\
Gride--Profiled & $-2.8\pm0.4$ & $-2.2\pm0.9$ & $-0.5\pm0.7$ & $+3.5\pm1.0$ & $+10.6\pm0.7$ \\
\bottomrule
\end{tabular}

\end{table}

Gride--Full overtakes MiND--Full between $\eta=0.05$ and $0.1$ and retains its advantage through the largest evaluated noise level, as its wider neighbor orders spread the added displacement over a longer baseline. Excluding the seven
benchmark manifolds with $d>5$ whose ambient dimension caps the candidate
search at $D\leq d+1$ preserves this ordering at $\eta=0.4$ (restricted-subset row of Table~\ref{tab:relative-noise-supp}: Gride--Full $23.4\%$ against
MiND--Full $37.6\%$ MPE).

 The lower panel of Table~\ref{tab:relative-noise-supp} gives the direction of this effect on the 14-manifold
complete-case subset on which all eight objectives are defined
(Section~\ref{app:component-results}): the signed mean error of every
objective rises monotonically with $\eta$, and all eight are positive at
$\eta=0.4$, so noise inflates the estimates. The inflation is strongest for Full angle-only,
from $+10.9\%$ at $\eta=0$ to $+32.9\%$ at $\eta=0.4$, and at $\eta=0.4$ each Gride objective is less inflated than its MiND counterpart (Gride--Full $+12.8\%$ against $+14.7\%$ for
MiND--Full, and Gride distance-only $+8.7\%$ against $+12.0\%$ for MiND
distance-only). The direction matches the short-scale noise inflation
described by Denti et al.~\cite{denti2022generalized}, and the Gride ordering
reflects the larger scale at which its ratio is read.

\FloatBarrier

\subsection{Profiling and distance drift under amplitude heterogeneity}
\label{app:gsm-results}

Table~\ref{tab:gsm-results-supp} gives all eight objectives, and
Table~\ref{tab:gsm-sweep-supp} shows how the three MiND objectives change as
amplitude heterogeneity increases across the image-class range.
At the focal level $\sigma_S=0.25$, no candidate reference mean direction lies below the observed one, so the angular-location penalty of Section~\ref{app:profiled-divergence} is paid at every candidate. Across five replicates its mean rises from $1.28$ to $14.76$ nats (KL divergence in natural-logarithm units) over $m=6,\ldots,100$, increasing at $87.4$ of the $94$ candidate steps on average; this rising penalty is what pulls the Full minimum below the distance evidence in Table~\ref{tab:gsm-sweep-supp}.

\begin{table}[H]
\centering
\caption{Eight calibrated objectives on amplitude-heterogeneous GSM samples at
$d=70$ in $D=100$, $\sigma_S=0.25$, over 30 replicates. Entries are mean $\pm$
standard deviation.}
\label{tab:gsm-results-supp}
\small
\begin{tabular}{lrr}
\toprule
Objective & Estimated ID & MPE (\%) \\
\midrule
MiND distance-only & $71.20\pm1.75$ & $2.48\pm1.72$ \\
Gride distance-only & $64.63\pm1.59$ & $7.67\pm2.27$ \\
Full angle-only & $6.26\pm0.03$ & $91.06\pm0.05$ \\
Profiled angle-only & $56.07\pm3.26$ & $19.90\pm4.65$ \\
MiND--Full & $22.80\pm1.76$ & $67.43\pm2.51$ \\
MiND--Profiled & $66.67\pm1.81$ & $4.76\pm2.58$ \\
Gride--Full & $17.22\pm1.64$ & $75.40\pm2.34$ \\
Gride--Profiled & $61.17\pm2.31$ & $12.62\pm3.29$ \\
\bottomrule
\end{tabular}
\end{table}

\begin{table}[H]
\centering
\caption{Observed mean direction $\hat\nu$ and MiND estimates across the GSM amplitude sweep ($d=70$ in $D=100$;
30 replicates). Entries are mean $\pm$ standard deviation. The known-amplitude control at $\sigma_S=0.25$ reuses the latent draws of the $\sigma_S=0$ level, so its values are those of the $\sigma_S=0$ row.}
\label{tab:gsm-sweep-supp}
\small
\begin{tabular}{rrrrr}
\toprule
$\sigma_S$ & $\hat\nu$ & Distance-only & Full & Profiled \\
\midrule
$0$ & $1.137\pm0.003$ & $65.07\pm1.01$ & $71.90\pm3.30$ & $64.37\pm0.89$ \\
$0.1$ & $1.055\pm0.008$ & $64.17\pm1.31$ & $52.05\pm3.23$ & $62.83\pm1.09$ \\
$0.15$ & $0.980\pm0.010$ & $64.60\pm1.30$ & $35.87\pm2.66$ & $62.47\pm1.25$ \\
$0.2$ & $0.903\pm0.012$ & $66.93\pm1.14$ & $28.49\pm2.50$ & $63.83\pm1.55$ \\
$0.25$ & $0.832\pm0.014$ & $71.20\pm1.75$ & $22.80\pm1.76$ & $66.67\pm1.81$ \\
$0.3$ & $0.769\pm0.014$ & $76.97\pm1.96$ & $19.95\pm1.55$ & $70.20\pm1.88$ \\
$0.35$ & $0.715\pm0.014$ & $84.47\pm2.16$ & $18.08\pm1.33$ & $73.70\pm2.49$ \\
\bottomrule
\end{tabular}

\end{table}

 As amplitude heterogeneity rises, Full moves away from the
distance-only result while Profiled follows it, because profiling removes
the angular-location penalty.  In this controlled setting, profiling returns the combined estimate to the distance evidence, and Profiled then reflects only the distance and concentration discrepancies. The distance-only column of Table~\ref{tab:gsm-sweep-supp} also records an upward drift, from $65.07$ to $84.47$, produced by the same preferential selection of low-amplitude neighbors (Main Section 3.4 and Corollary~\ref{cor:supp-gsm-shell}).

\FloatBarrier

\subsection{Angular-reference transfer and normalization on images}
\label{app:image-diagnostics}

\begin{figure}[ht]
\centering
\includegraphics[width=0.95\linewidth]{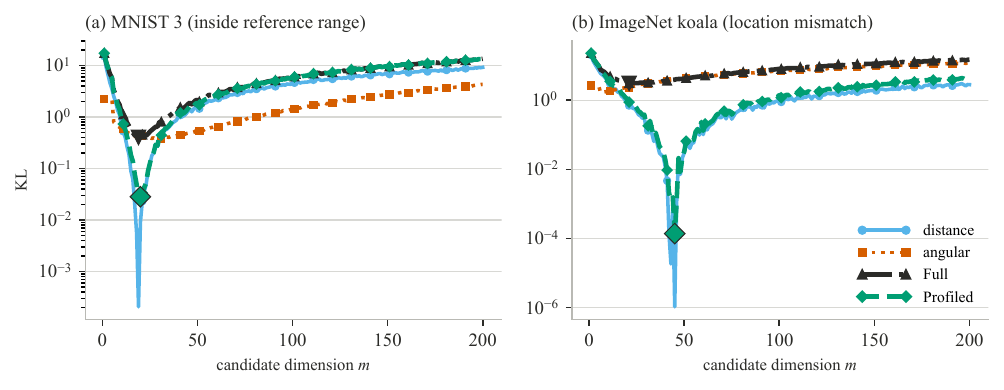}
\caption{Complete KL analysis for MNIST digit 3 and ImageNet koala using one
fixed reference replicate and the MiND distance model. The legend distinguishes components by color, line
style, and marker; enlarged symbols mark the Full and Profiled joint minima.}
\label{fig:kl-diagnosis-supp}
\end{figure}

 The raw CIFAR-10 and ImageNet mean directions lie below their reference ranges, and a Full--Profiled separation accompanies the gap. Main Table 2 gives the raw
statistics and estimates; Figure~\ref{fig:kl-diagnosis-supp} shows the
underlying objectives. For the fixed koala reference, angular discrepancy
increases with candidate dimension (Spearman correlation $0.998$), consistent
with the angular-location penalty of Main Section 3.1, which profiling
removes exactly. The Full angular component reaches its minimum at
$m=6$, the Full objective at $m=21$, and both the distance component and the Profiled
objective at $m=45$. The two panels connect these minima to the reference ranges
in Table~\ref{tab:reference-range-supp}.

\paragraph{Normalization controls and per-center geometry}
\label{app:image-controls}

\begin{figure}[ht]
\centering
\includegraphics[width=0.86\linewidth]{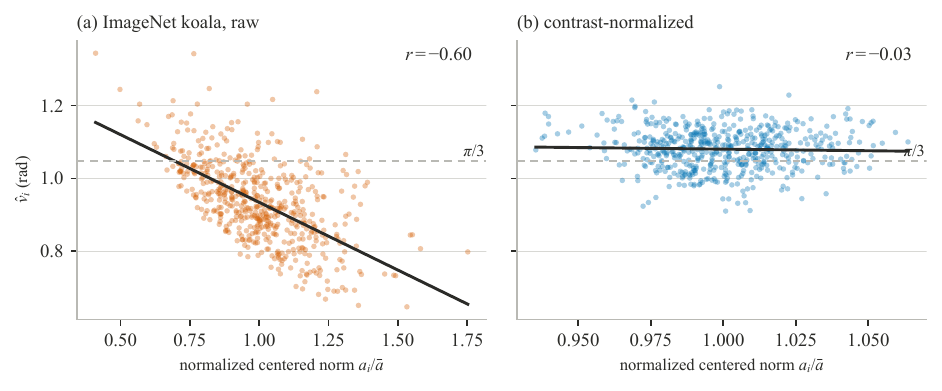}
\caption{Per-center mean direction against normalized centered norm $a_i/\bar a$ for raw and contrast-normalized ImageNet koala data. Normalization largely removes the
broad norm tail and weakens the negative relation. Solid lines are least-squares trends; the
dashed horizontal line marks $\pi/3$, and $r$ denotes the Pearson correlation
between centered norm and per-center mean direction.}
\label{fig:nu-scatter-supp}
\end{figure}

Reducing sample-amplitude variation moves the observed mean direction toward
the reference range and narrows the Full--Profiled separation.
Table~\ref{tab:diagnostics-supp} reports the full normalization results, and
Figure~\ref{fig:nu-scatter-supp} shows the center-level relation for ImageNet
koala.

In the generative model of Section S1.2 the
amplitude $S_i$ multiplies every coordinate of observation $i$, and per-image
contrast normalization divides by exactly such a per-observation scale: the
Gaussian scale mixture reading of natural images treats contrast as the
per-observation scale variable. Coordinate-wise standardization or range
scaling instead applies a common translation and one diagonal rescaling to all observations. With equal scaling factors across coordinates this reduces to a single global rescaling, which
leaves the coefficient of variation of centered norm, the neighbor selection,
and the mean direction unchanged.

\begin{table}[H]
\centering
\caption{Normalization controls. The observed-data mean direction and Full
and Profiled estimates are reported for raw, centered-radial, and
contrast-normalized data over five reference replicates. Full and Profiled
denote MiND--Full and MiND--Profiled.}
\label{tab:diagnostics-supp}
\small
\setlength{\tabcolsep}{3.5pt}
\begin{tabular}{l rrr rrr rrr}
\toprule
 & \multicolumn{3}{c}{raw} & \multicolumn{3}{c}{centered radial} & \multicolumn{3}{c}{contrast-norm.} \\
\cmidrule(lr){2-4}\cmidrule(lr){5-7}\cmidrule(lr){8-10}
Dataset & $\hat\nu$ & Full & Profiled & $\hat\nu$ & Full & Profiled & $\hat\nu$ & Full & Profiled \\
\midrule
MNIST 3 & 1.166 & 20.4 & 20.2 & 1.198 & 20.2 & 20.0 & 1.135 & 20.8 & 20.8 \\
MNIST 7 & 1.157 & 14.4 & 14.8 & 1.201 & 14.8 & 14.8 & 1.143 & 14.8 & 14.8 \\
CIFAR-10 bird & 0.898 & 18.8 & 34.0 & 0.982 & 22.5 & 33.0 & 1.064 & 28.2 & 34.0 \\
CIFAR-10 cat & 0.946 & 21.2 & 32.8 & 1.030 & 24.4 & 32.8 & 1.087 & 30.6 & 35.4 \\
ImageNet koala & 0.934 & 24.9 & 45.0 & 1.028 & 40.6 & 53.4 & 1.080 & 72.2 & 70.0 \\
ImageNet butterfly & 0.915 & 28.1 & 58.8 & 1.027 & 60.3 & 74.8 & 1.040 & 85.2 & 104.0 \\
\bottomrule
\end{tabular}

\end{table}

For CIFAR-10 and ImageNet, both normalizations raise $\hat\nu$ toward the
reference range and narrow the Full--Profiled separation. MNIST begins inside its reference range, and its Full and Profiled estimates stay close under all transformations. Gride--Profiled shows the same upward shift under contrast normalization.

The centered-norm measure uses $\operatorname{CV}(a)$ defined in Section~\ref{app:design-gsm}.
Raw koala and butterfly have Pearson correlations of $-0.60$ between centered norm and per-center mean direction, while contrast normalization reduces the koala correlation to $-0.03$. Their centered-norm variation, $\operatorname{CV}(a)=0.15$--$0.18$, is also much larger than that of a uniform 70-ball sample ($0.014$). Together with the GSM control, these observations support amplitude variation as a substantial contributor to the image angular shift.

\paragraph{Additional image classes}
\label{app:additional-real}

The class survey examines whether low angular location extends beyond the
examples analyzed in the Main paper.

\begin{table}[H]
\centering
\caption{Per-class angular statistics at $k=10$. Here $r$ is the Pearson
correlation between per-center mean direction and centered norm, and
$\operatorname{CV}(a)$ is the population standard deviation of
$a_i=\|x_i-\bar x\|_2$ divided by their mean, with $\bar x$ the class mean.
ImageNet class codes are WordNet synset identifiers.}
\label{tab:class-survey}
\small
\begin{tabular}{ll rr rrr}
\toprule
Source & Class & $N$ & $D$ & $\hat\nu$ & $\operatorname{CV}(a)$ & $r(\hat\nu_i, \|x_i-\bar x\|)$ \\
\midrule
CIFAR-10 & airplane & 5000 & 3072 & 0.939 & 0.291 & -0.08 \\
CIFAR-10 & automobile & 5000 & 3072 & 1.037 & 0.219 & -0.04 \\
CIFAR-10 & bird & 5000 & 3072 & 0.898 & 0.330 & -0.10 \\
CIFAR-10 & cat & 5000 & 3072 & 0.946 & 0.256 & -0.08 \\
CIFAR-10 & deer & 5000 & 3072 & 0.903 & 0.312 & -0.13 \\
CIFAR-10 & dog & 5000 & 3072 & 0.938 & 0.243 & -0.10 \\
CIFAR-10 & frog & 5000 & 3072 & 0.917 & 0.320 & 0.08 \\
CIFAR-10 & horse & 5000 & 3072 & 0.986 & 0.240 & -0.25 \\
CIFAR-10 & ship & 5000 & 3072 & 0.936 & 0.279 & -0.08 \\
CIFAR-10 & truck & 5000 & 3072 & 1.022 & 0.196 & -0.33 \\
ImageNet & n01882714 & 547 & 150528 & 0.934 & 0.180 & -0.60 \\
ImageNet & n02086240 & 548 & 150528 & 0.940 & 0.200 & -0.50 \\
ImageNet & n02087394 & 947 & 150528 & 0.900 & 0.222 & -0.39 \\
ImageNet & n02094433 & 766 & 150528 & 0.982 & 0.196 & -0.53 \\
ImageNet & n02100583 & 826 & 150528 & 0.892 & 0.240 & -0.33 \\
ImageNet & n02100735 & 614 & 150528 & 0.912 & 0.201 & -0.15 \\
ImageNet & n02279972 & 648 & 150528 & 0.915 & 0.149 & -0.60 \\
\bottomrule
\end{tabular}

\end{table}

Fifteen of the 17 classes in Table~\ref{tab:class-survey} have $\hat\nu<1.0$,
and within CIFAR-10 the two exceptions also have the smallest centered-norm spread, consistent with an amplitude effect on angular location.

\FloatBarrier

\subsection{Layer-wise CNN profiles}
\label{app:cnn-diagnostics}

Table~\ref{tab:cnn-peaks-supp} reports, for each architecture, the interior
peak of the median Gride--Profiled curve together with the largest magnitude
of the median paired difference
between Gride--Full and Gride--Profiled; that largest difference coincides with the peak in every architecture. Relative depth follows the definition of
Section~\ref{app:design-cnn}. On AlexNet, the TWO-NN medians over the seven categories of the
first subsample at the input, the three max-pooling checkpoints, and the three
linear layers ($29.7$, $69.1$, $65.3$, $38.6$, $28.0$, $21.7$, $17.2$)
closely reproduce the class-averaged profile reported by Ansuini et
al.~\cite[Fig.~3A]{ansuini2019intrinsic}. Layer medians of the observed mean direction $\hat\nu$ range from $0.780$ to $1.106$ over the four networks: they lie at $0.78$--$0.87$ at the first pooling checkpoint of each network, rise through depth, and reach $1.04$--$1.11$ at the output layer. There the median paired Full$-$Profiled difference is at most six dimensions (Table~\ref{tab:cnn-peaks-supp}, lower panel).

\begin{table}[H]
\centering
\caption{Peak of the median Gride--Profiled curve across seven categories and
three subsamples. Full and Profiled denote Gride--Full and Gride--Profiled, and
the largest absolute difference between them occurs at the same layer in each
architecture. The last column is the median of the per-cell paired
Full$-$Profiled differences across category--subsample cells, not the
difference of the two reported medians. Layer names denote mathematical checkpoints
along the sequence of network transformations. Lower panel: mean direction at the first pooling checkpoint and at the output layer, with the output medians of both objectives and their paired difference.}
\label{tab:cnn-peaks-supp}
\small
\setlength{\tabcolsep}{2pt}
\begin{tabular}{llrrrrr}
\toprule
Architecture & Peak layer & depth (\%) & $\hat\nu$ & Full & Profiled & Median paired Full$-$Profiled \\
\midrule
AlexNet & second max-pooling checkpoint & 22.2 & 0.800 & 17 & 115 & -100 \\
VGG-16 & second max-pooling checkpoint & 23.5 & 0.802 & 18 & 197 & -179 \\
ResNet-18 & second residual stage & 47.4 & 0.916 & 37 & 248 & -214 \\
ResNet-34 & second residual stage & 42.9 & 0.919 & 37 & 245 & -208 \\
\bottomrule
\end{tabular}
\\[6pt]
\begin{tabular}{lrrrrr}
\toprule
Architecture & $\hat\nu$ first pooling & $\hat\nu$ output & Full output & Profiled output & Median paired Full$-$Profiled \\
\midrule
AlexNet & 0.801 & 1.106 & 22 & 22 & -1 \\
VGG-16 & 0.780 & 1.106 & 18 & 20 & -2 \\
ResNet-18 & 0.873 & 1.069 & 29 & 33 & -4 \\
ResNet-34 & 0.855 & 1.038 & 24 & 31 & -6 \\
\bottomrule
\end{tabular}

\end{table}

\FloatBarrier
\bibliographystyle{elsarticle-num}
\bibliography{refs}